\documentclass[twocolumn,twoside,journal]{IEEEtran}

\usepackage{pdfsync}
\usepackage{graphicx}
\usepackage{color}
\usepackage{verbatim}
\usepackage{pseudocode}
\usepackage{cite}
\usepackage{url}
\usepackage{mathrsfs}
\usepackage{rsfso}
\usepackage{labtex}
\usepackage{inkscape}
\usepackage[font=footnotesize]{caption,subfig}
\usepackage{dblfloatfix}
\usepackage{flushend}

\graphicspath{{./figs/}}

\let\euclinner\inner
\renewcommand{\inner}[2]{\ensuremath{\euclinner{#1}{#2}_g}}
\let\euclnorm\norm
\renewcommand{\norm}[1]{\ensuremath{\euclnorm{#1}_g}}

\def\FF{\bbold{F}}

\def\FFp{\ensuremath{\FF^+}}
\def\FFm{\ensuremath{\FF^-}}

\def\vv{\ensuremath{\mathbf{v}}}
\def\uu{\ensuremath{\mathbf{u}}}
\def\ww{\ensuremath{\mathbf{w}}}
\def\rr{\ensuremath{\mathbf{r}}}
\def\pp{\ensuremath{\mathbf{p}}}
\def\PT{\mathcal{P}_\bbold{T}}
\def\PN{\mathcal{P}_\bbold{N}}
\def\PP{\mathcal{P}}

\begin{document}
\pdfmapfile{+rsfso.map}

\title{A propagative model of simultaneous 
impact: existence, uniqueness, and design consequences}
\author{Vlad~Seghete,~\IEEEmembership{Student~Member,~IEEE,}
Todd~D.~Murphey,~\IEEEmembership{Member,~IEEE}%
\thanks{V. Seghete and T. Murphey are with the Department of Mechanical Engineering
at Northwestern University, 2145 Sheridan Rd., Evanston, IL 60660.}
}

\date{}
\maketitle

\begin{abstract}
  This paper presents existence and uniqueness results for a
  propagative model of simultaneous impacts that is guaranteed to conserve
  energy and momentum in the case of elastic impacts with extensions to
  perfectly plastic and inelastic impacts. A
  corresponding time-stepping algorithm that guarantees conservation of
  continuous energy and discrete momentum is developed, also with extensions to
  plastic and inelastic impacts. The model is illustrated in simulation using
  billiard balls and a two-dimensional legged robot as examples; the
  latter is optimized over geometry and gait parameters to achieve unique
  simultaneous impacts.\\

\indent\textit{Note to Practitioners}---Simultaneous
impacts are a common occurrence in manufacturing and robotic
applications. Simulation-based techniques predicting the motion of
a mechanical system subject to simultaneous impacts often use numerical
routines that make algorithmic assumptions about impact in order
to make the simulation more tractable. Such assumptions have the
potential to significantly influence the simulation outcome and
may even invalidate results.  This paper provides tools for simulating
simultaneous impacts, verifying a simulation that deals with
simultaneous impacts, and designing a system so that the simulation
will be less dynamically sensitive to indeterminacy in simultaneous
impact.
\end{abstract}


\IEEEpeerreviewmaketitle

\pagestyle{plain}

\section{Introduction}
\label{sec:intro}
\begin{figure}[b]
  \vspace{-1em}
      \centering
      \includegraphics[width=\columnwidth]{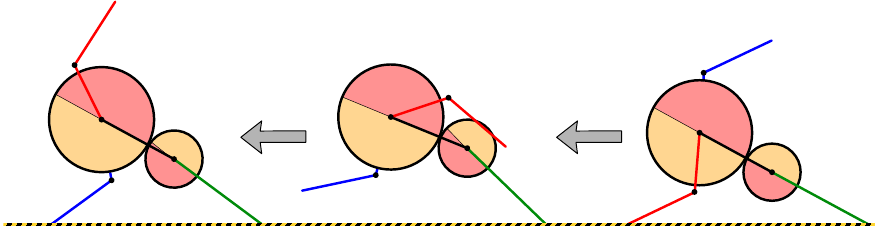}
      \caption{The running robot model (TREX) that we used in order to
      illustrate the results of this paper. The arrow depicts the intended
      direction of movement which results from a counter-clockwise
      torque applied to the legs at the hips.}
      \label{fig:trex_intro}
\end{figure}

\IEEEPARstart{I}{n} this paper we investigate a propagative impact
model for rigid body
systems\cite{baraff_analytical_1989,chatterjee_new_1998} and the
conditions under which it provides a deterministic prediction for
the outcome of simultaneous impacts.  Our motivating system is
pictured in Fig.~\ref{fig:trex_intro} and represents a simplified
model based on the geometry of numerous legged robots
\cite{eich_versatile_2008,quinn_parallel_2003,jeans_impass:_2009,
  lyons_rotational_2005,saranli_rhex:_2001}
While contact
modeling has received significant attention in relation to plastic
impacts and established
contacts\cite{xiong_algebraic_2007,staffetti_analysis_2009}, little
work has been done to address the issue of simultaneous non-plastic
impacts. Such interactions naturally occur in the system of
Fig.~\ref{fig:trex_intro}---between a foot and the floor while the
tail is still in contact---but also in the gait of robots such as
RHex\cite{saranli_rhex:_2001} or IMPASS\cite{jeans_impass:_2009}.
As simulation has proven a strategic
tool in the design of mobile robots \cite{primerano_case_2011}, we
present theoretical results that can be used both as building blocks
for physical simulations as well as a validation tool for existing
simulation methods.

When simulating impacts in physical systems two main approaches have
been widespread. The most popular method is to solve unilateral
contact problems implicitly---usually by solving a linear complementarity
problem (LCP) associated with every time step of the simulation\cite{
berard_davinci_2007,stewart_implicit_2000,chakraborty_implicit_2007,
trinkle_dynamic_1997, moreau_numerical_1996,moreau_numerical_1999,
tong_liu_computation_2005,_newton_????}.  The defining feature 
of this implicit approach is that all the collisions detected over a
time step are processed together with the dynamics for that time step, which
eases implementation and scalability. In addition, implicit LCP
methods potentially have very fast execution times and are 
parallelizable\cite{_physx_????, _ode:_????,_bullet_2011}. This makes such
methods ideal in most real-time physical simulations---e.g. interactive
graphics applications, video games, etc. However, the implicit nature of such
algorithms is double-edged: at the cost of a high-performing low-hassle
implementation they can allow for unrealistic behavior---especially noticeable when
dealing with non-plastic impacts.
\begin{figure*}
      \vspace{-2em}
      \centering
      \subfloat[Three spheres restricted to move on a line. The 
      configuration space is three dimensional,
      with $x_A$, $x_B$ and $x_C$ as the configuration variables.
      This system is a simplification of
      the Newton's cradle toy.]{
        \label{fig:cradle}
        \includegraphics{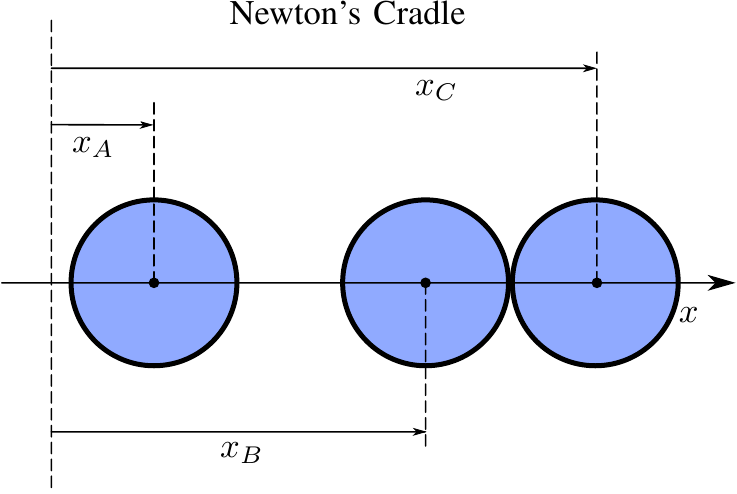}
      }
      \qquad\quad
      \subfloat[Energy loss versus initial velocity for the simplified
      Newton's cradle simulated in Bullet (dots) and using our method (dotted
      lines). Each band of points and dotted line represents one of three
      coefficients of restitution (COR) tested: 1, 0.7 and 0.]{
        \label{fig:bullet}
        \includegraphics{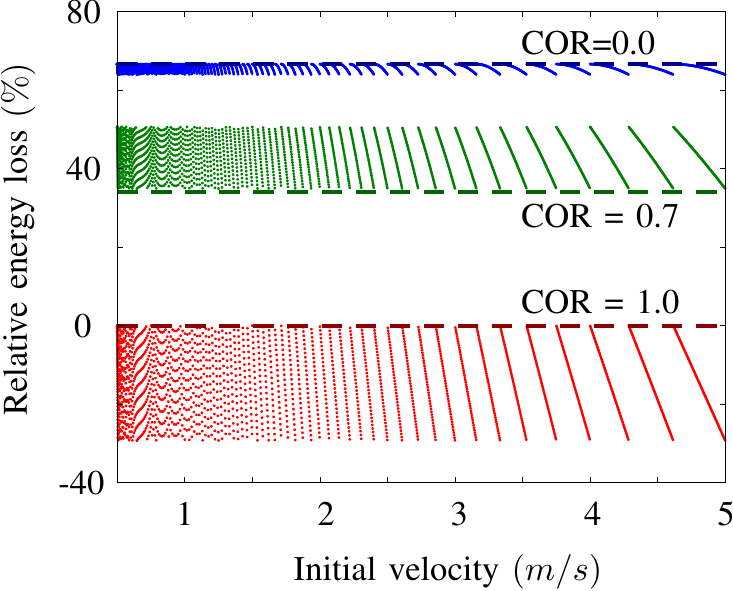}
      }
      \caption{Simplified Newton's cradle showing unrealistic behavior in the Bullet
      physics engine. Notice the
      inconsistencies in energy behavior present in the Bullet data points.}
      \label{fig:bad_bullet}
      \vspace{-0.2in}
\end{figure*}
To illustrate such behavior, we simulated the three sphere system shown in
Fig.~\ref{fig:cradle} using the popular Bullet physics
engine\cite{_bullet_2011}.  We let spheres $B$ and $C$ be in contact initially
and had sphere $A$ impact $B$ with varying velocities.
Figure~\ref{fig:bullet} shows the amount of energy lost through this impact for
three different values of the coefficient of restitution. For 
perfectly elastic collisions the system energy increased by as much as $30\%$
due to the impact.  Furthermore, this error does not scale with the
time step; while varying the time step frequency between 60~Hz and 6~kHz the
error did not change, indicating we cannot
expect this algorithm to converge to a ``true'' solution as the step size
goes to zero. Aditionally, even at velocities for
which the energy was conserved, the solution did not come close to the
experimental results\cite{donahue_newtons_2008}.
It is worth noting here that while most LCP methods are implicit,
work has been done that incorporates the LCP approach to impacts into
event-driven integration schemes \cite{mirtich_impulse_1996,hahn_realistic_1988}
as discussed below.

An alternative method to modeling rigid body collisions is to treat
impacts as impulsive events where a jump in velocity occurs at the
moment of impact.  At each impact a reset map determines
the post-impact state of the system. The main benefit of this
approach is the explicit nature of the reset map which allows one
control of the simulation's physical accuracy and compliance with
energy conservation and restitution laws at the time of impact.
Using an explicit method also gives us the freedom to employ a
\emph{propagative} model for handling simultaneous impacts, a model
which has better agreement with experimental
data\cite{donahue_newtons_2008} than the implicit methods used in obtaining
the results of Fig.~\ref{fig:bullet}.
Propagative collision models solve
simultaneous  impacts sequentially, splitting one simultaneous
impact into as many two body impacts as needed to reach a solution.
The method is based on assuming an infinitesimal gap between all
contacts and solving for two-body impacts in an order of one's
choice. While there is experimental evidence that
better qualitative results can be achieved by using soft-body
models\cite{jia_simultaneous_2011,siciliano_state_2009} and similar
approaches
\cite{zhao_energy_2008,liu_frictionless_2008,liu_frictionless_2009}, the
errors incurred by using a rigid body propagative model are slight
at best\cite{donahue_newtons_2008} and certainly not to the degree
of the errors presented in Fig.~\ref{fig:bullet}.\footnote{Note
that the results presented in \cite{donahue_newtons_2008} reflect
the behavior of the system when the initial velocities have relatively
high values compared to the stiffness of the bodies involved in the
collisions.} The rigid body approach we undertake allows us to model
more complex systems, like the legged robot model shown in
Sec.~\ref{sec:design}.  Unfortunately, the main drawback in using
rigid body propagative approaches, as discussed in
\cite{baraff_analytical_1989} and \cite{liu_frictionless_2008}, is
the inadequate modeling of continuum mechanics phenomena. In particular,
the coupling effects between simultaneous impacts cannot be generally
addressed.  In the rigid body limit such coupling
effects manifest themselves as a lack of unique solutions to the equations
governing simultaneous impact.  However, rather than proposing to
solve the general simultaneous impact problem, we focus instead on
identifying cases in which coupling effects do not generate
non-uniqueness in the rigid body limit for simultaneous impacts.
Such cases, therefore, can be modeled under the rigid body
assumption provided mechanism
and gait design goals are set appropriately.

As major contributions the present paper provides the following:
\begin{enumerate}
  \item sufficient conditions for solution existence for the propagative model
    of elastic impact, assuming two simultaneous impacts
  \item sufficient conditions for solution uniqueness for the propagative model
    of elastic impact, assuming $n$ simultaneous impacts; when uniqueness is
    satisfied, existence is guaranteed for arbitrary numbers of simultaneous
    impacts.
  \item a time stepping algorithm which preserves discrete momentum and continuous
    energy
  \item extensions of 1-3 above to the cases of plastic and inelastic impacts,
    as well as automated distinction between plastic and non-plastic impacts
    when using the time-stepping method
  \item application to three example systems: Newton's cradle, billiards, and
    a two dimensional legged robot model robot which includes both external
    forcing and Coulomb friction
\end{enumerate}

The rest of the paper is structured around the above contributions.
Section~\ref{sec:single_collision} goes over
the collision model in the simple case of one isolated impact and
introduces notation that will be used throughout the rest of the paper.
Section~\ref{sec:simultaneous_collisions} expands the basic collision model by
extending it to the case of simultaneous collisions and explains its use in the
example case of Newton's cradle.  In Sec.~\ref{sec:existence} we examine
existence arguments for the solutions of our algorithm, giving an upper bound
on the number of iterations needed to achieve a feasible result for the case of
two simultaneous contact points. The existence results shown here correspond to
and extend those presented in\cite{seghete_variational_2010}, substituting a
rigorous mathematical proof in place of a geometric argument.  In
Sec.~\ref{sec:uniqueness} we investigate the uniqueness of the solutions
obtained by the propagative method, for both the case of $n$ perfectly elastic
and perfectly plastic simultaneous impacts. For the plastic case the solution is
trivially unique. However, for the elastic case we find that, while in certain
cases, including Newton's cradle, our approach gives unique solutions,
in general we have a non-unique, although finite, number of outcomes.
Similarly, standard implicit
methods do not provide uniqueness results for their solutions, as the
final solution is usually dependent on both the initial condition and the particular
solver. Our approach has
the advantage of offering a set of countably many solutions, all of which
satisfy the LCP criteria. 
Moreover, the non-unique solutions that we present have been posited to
correspond to uncertainty in the elastic body physical
system\cite{chatterjee_new_1998}.  
This gives rise to an interesting question: could we leverage
control of the configuration at impact to always obtain predictable collision
outcomes when dealing with simultaneous impacts?  In Sec.~\ref{sec:design} we
give an affirmative answer to this question---an 
extension to our work in\cite{seghete_conditions_2012}---and present its use
on two example systems: an extension of the simplified system in
Fig.~\ref{fig:cradle} and a two-legged and tailed robot intended for locomotion
seen in Fig.~\ref{fig:trex_intro}. All the simulation results are based on a time
stepping scheme consisting of variational integrators which is briefly
described in Sec.~\ref{sec:discrete} and the foundations of which were
discussed in more detail in\cite{seghete_multiple_2009}
and\cite{seghete_variational_2010}. 

\section{Basic Impact Model}
\label{sec:single_collision}
The following presents well known results concerning the equations
governing impacts treated with an impulsive approach. We present this
both for reference and to introduce the reader to notation. Of 
particular note is our definition of inner product and norm on
the cotangent space---see \eqref{eq:dualmetric}---which makes use of 
the dual of the kinetic energy metric on the tangent space.

Assume a simple mechanical system described by configuration $q$ and
Lagrangian
 $ L(q,\dot{q})=K(q,\dot{q}) - V(q,\dot{q})$.
The equations of motion for such a system can be derived by applying
a variational principle, and are the known Euler-Lagrange
equations\cite{marsden_introduction_1999}:
\begin{equation}
  \label{eq:euler-lagrange}
  \derv{}{t}\partl{L}{\dot{q}}(q,\dot{q})-\partl{L}{q}(q,\dot{q})=0.
\end{equation}
For a collision at time $t_*$ we say that the velocity at that time is
\emph{infeasible} if it points ``into'' the contact manifold---opposite 
to the contact manifold normal:
\begin{equation}
  \D\phi(q_*)\cdot\dot{q}_*  < 0,
  \label{eq:infeasibility}
\end{equation}
where $q_* = q(t_*)\in Q$ is the configuration at time of impact and
$\phi:Q\to\reals$ is the gap function describing the contact manifold
$\mathcal{C}\subset Q$\cite{moreau_numerical_1999}---it takes positive values
in the feasible region of space, negative values in the infeasible region, and
is identically zero when $q$ is on the contact manifold.  $\D\phi$ is the
first derivative of the gap function with respect to its argument, and hence
is a covector that belongs to the cotangent space at $q_*$:
 $ \D\phi(q_*)=\partl{\phi}{q}(q_*) \in T_{q_*}^*Q$,
and provides the normal to the contact manifold. 
Using a variational approach one
finds that the equations governing a single elastic collision at time $t_*$ are:
\begin{subequations}
\label{eq:one_impact}  
\begin{gather}
	\frac{\partial L}{\partial \dot{q}} \Big |_{t_*^-}^{t_*^+} 
    = \lambda \D\phi(q_*), \label{eq:el_momentum}\\
	\left[ \frac{\partial L}{\partial \dot{q}} \cdot \dot{q} 
    - L\right]_{t_*^-}^{t_*^+} = 0 \label{eq:el_energy}.
\end{gather}
\end{subequations}
Here $\lambda$ is a
Lagrange multiplier such that $\lambda\D\phi$ 
can be interpreted as the impulse imparted to the system through impact. 
The two equations have
a classical interpretation: \eqref{eq:el_momentum} is the conservation of 
momentum---tangentially to the impact
manifold---and \eqref{eq:el_energy} is the conservation of energy through the impact.

From here on, we work under the assumption that we are dealing with
a non-degenerate simple mechanical system such that 
potentials do not depend on the
velocity $\dot{q}$, and the kinetic energy term is quadratic in
$\dot{q}$ . Under this assumption, we can write the
Lagrangian as
$	L(q,\dot{q}) = \tfrac{1}{2}\dot{q}\tr  M(q) \dot{q} - V(q)$,
where $M(q)$ is the mass matrix
$M(q) = \partial_{\dot{q} \dot{q}}L(q,\dot{q})$
and is positive definite.
The system in \eqref{eq:one_impact} becomes 
\begin{align*}
	\dot{q}\tr(t_*^+) M - \dot{q}\tr(t_*^-) M &= \lambda_*\D\phi_* , \\
    \dot{q}\tr(t_*^+) M \dot{q}(t_*^+) &= \dot{q}\tr(t_*^-) M (t_*^-) ,
\end{align*}
where, for ease of notation, we dropped the $q$-dependency of $M$ and
$\phi_*$, implicitly assuming they are 
evaluated at $q_*$, the impact configuration.
We can rewrite these equations as
\begin{subequations}
    \label{eq:cont_nrg_conserv}
\begin{gather}
	\pp^+ = \pp^- + \lambda \uu, \\
    \pp^+ M^{-1}\pp^{+\mathrm{T}} = \pp^- M^{-1}\pp^{-\mathrm{T}},
\end{gather}
\end{subequations}
where $\pp^\pm = \dot{q}_{*^\pm}\tr M$ is the momentum before and after the
collision and $\uu = \D\phi_*\tr$ is the normal to the manifold of impact (for
the purposes of this paper we use bold notation to denote covectors, which in
our case are elements of the cotangent bundle---e.g.  $\pp,\uu \in
T_{q_*}^*Q$---while regular script denotes vectors: $q \in Q$, $\dot{q} \in
T_{q_*}Q$, etc.) Also, for the remainder of the paper, the norms and dot
products between covectors are assumed to be those defined under the local
kinetic energy metric\cite{bullo_section_2010}:
\begin{subequations}
  \label{eq:dualmetric}
\begin{align}
  \inner{\uu}{\vv} &= \uu M^{-1} \vv\tr, \\
  \norm{\uu}^2 &= \left<\uu,\uu\right>.
\end{align}
\end{subequations}
Using this notation, we redefine infeasibility in terms of 
momentum. We say that $\pp^-$ is \emph{infeasible} with respect
to the contact manifold represented by the normal $\uu$ if
\begin{equation}
  \inner{\pp^-}{\uu} < 0.
  \label{eq:momentum_infeasibility}
\end{equation}
By the same token, $\pp^-$ is \emph{feasible} if $\inner{\pp^-}{\uu} \ge 0$.

Assuming an infeasible $\pp^-$, we solve for $\pp^+$ using
\eqref{eq:cont_nrg_conserv}:
\begin{subequations}
  \label{eq:Gamma}
\begin{align} 
  	\pp^+ &= \pp^- \Gamma(\uu),\label{eq:momentum_map} \\
    \Gamma(\uu) &= I - 2M^{-1}\frac{\uu\tr\uu}{\norm{\uu}^2} = I - 2\frac{M^{-1} \uu\tr \uu}{\uu M^{-1} \uu\tr}.
\end{align}
\end{subequations}
Here $\Gamma(\uu)$ is a momentum (reset) map\cite{chatterjee_new_1998}
that describes
an instantaneous change in momentum due to an impact with 
a manifold normal $\uu$.

Equation \eqref{eq:momentum_map} is only one of the two
solutions to the system in \eqref{eq:cont_nrg_conserv}, but
is the only feasible one. Indeed, after
eliminating $\pp^+$ we are left with a quadratic equation in $\lambda$:\quad
$    \lambda^2 \norm{\uu}^2 + 2\lambda \inner{\pp^-}{\uu} = 0$.
The feasible value is given by
$  \lambda = -2\inner{\pp^-}{\uu}/ \norm{\uu}^2$.
Note that if we write \eqref{eq:infeasibility} using the notation
just introduced, we have that $\lambda >0$. This is consistent with the LCP
formulation of contact, as $\lambda>0$ satisfies the classic
complementarity conditions.

The mapping $\Gamma(\uu)$, as defined in (\ref{eq:momentum_map}), has several
properties that will be useful in Sec.~\ref{sec:simultaneous_collisions},
Sec.~\ref{sec:existence}, and Sec.~\ref{sec:uniqueness}. First, 
\begin{equation}
  \Gamma(\uu)^2 = \Gamma(\uu)\Gamma(\uu) = I,
  \label{eq:reversal}
\end{equation}
which means that resolving an impact across a manifold twice in a row
returns the original momentum. Also, $\Gamma$ is not dependent on the magnitude
of $\uu$, only on its direction: 
\begin{equation}
  \Gamma(\alpha\uu) = \Gamma(\uu).
  \label{eq:magninvariance}
\end{equation}

\section{Simultaneous Collisions}
\label{sec:simultaneous_collisions}
Simultaneous impact can occur if two contacts are made at the same time or 
if one contact is already present when a second impact occurs.
When an impact is assumed to be plastic
 the equations governing the interaction are
 \begin{subequations}
   \label{eq:el_plastic}
\begin{gather}\label{eq:simultaneous_plastic_impact}
	\frac{\partial L}{\partial \dot{q}} \Big |_{t_*^-}^{t_*^+} =
    \sum_{i\in\mathcal{U}\cup\mathcal{V}} \lambda_i \D\phi_i(q_*), \\
    \D\phi_i(q_*)\dot{q}_*^+  = 0,\quad \lambda_i \ge 0,
    \quad\forall i\in\mathcal{U}\cup\mathcal{V},
\end{gather}
\end{subequations}
where $\mathcal{U}$ is the set of indices of manifolds already in contact
while $\mathcal{V}$ is the set of indices of 
new contact manifolds.
These equations generate a unique solution,
which corresponds to eliminating the portion of $\dot{q}_*$ that is 
orthogonal---under the kinetic metric---to the manifolds of collision
$\phi$ at the time of impact.
Using the notation introduced in Sec.~\ref{sec:single_collision}, the outcome of 
a plastic impact is given by
\begin{equation}
  \pp^+ = \mathcal{P}_{\mathrm{null}\left(\mathrm{Span}\left\{\D\phi_i(q_*)
  \left|i\in\mathcal{U}\cup\mathcal{V}
  \right\}\right)\right.}\left(\pp^-\right),
  \label{eq:plastic_outcome}
\end{equation}
where $\mathcal{P}_S(\pp)$ represents the projection of covector $\pp$ onto a
subspace $S$. In essence, $\pp^+$ is always tangent to all
contact manifolds---or, equivalently, it is orthogonal to all contact manifold
normals at the current configuration.

The uniqueness of the result is lost when considering elastic impacts:
\begin{subequations}
  \label{eq:el_elastic}
\begin{gather}
	\frac{\partial L}{\partial \dot{q}} \Big |_{t_*^-}^{t_*^+} =
    \sum_{i\in\mathcal{U}\cup\mathcal{V}} \lambda_i \D\phi_i(q_*), 
    \label{eq:el_elastic_momentum}\\
   \left[ \partl{L}{\dot{q}} \cdot \dot{q} - L\right]_{t_*^-}^{t_*^+}= 0,\\
   \lambda_i \ge 0, \quad\forall i\in\mathcal{U}\cup\mathcal{V},
\end{gather}
\end{subequations}
These equations form a fully determined system only when there 
is a single term in the summation on the right hand side of
\eqref{eq:el_elastic_momentum}---in which case they reduce to the case
treated in Sec.~\ref{sec:single_collision}.
Otherwise, the equations governing the impact
dynamics become underdetermined (there are more variables than equations).
This gives rise to a whole continuum of solutions. For generating trajectories
in simulation, an element of this
continuum must be chosen. An a priori relation between lambdas could be chosen in
order to solve this dilemma. However, it is unclear what physical principle
to use. Instead, we expand on a version of the propagative method
discussed in\cite{chatterjee_new_1998}. The reason behind using a propagative
model of simultaneous impact is
twofold: the method gives unique and correct results in simple,
intuitive
cases in which other methods fail---e.g.  Newton's cradle---and it also
provides at most a finite number of valid solutions in other cases, as
discussed in Sec.~\ref{sec:uniqueness}. 

We investigate the simplest case of simultaneous impacts: that of two
manifolds of contact, such that $\mathcal{U}\cup\mathcal{V}=\{a,b\}$.
Suppose that the impact occurs across two manifolds at
the exact same time, such that \eqref{eq:el_elastic} becomes
\begin{subequations}
\label{eq:two_impact}
\begin{gather}
	\frac{\partial L}{\partial \dot{q}} \Big |_{t_*^-}^{t_*^+} =
	 \lambda_a \textrm{D}\phi_a(q_*) + \lambda_b \D\phi_b(q_*), \\
	\left[ \frac{\partial L}{\partial \dot{q}} \cdot \dot{q} - L\right]_{t_*^-}^{t_*^+} = 0,\\
    \lambda_a \ge 0,\quad\lambda_b\ge0,
\end{gather}
\end{subequations}
which has the same number of equations as (\ref{eq:one_impact}), but one extra
variable. Instead, a propagative approach consists of 
applying (\ref{eq:momentum_map}) repeatedly until a
feasible momentum is found---whether this is even operationally valid is
something we address in Sec.~\ref{sec:existence}. Using the notation of the
previous section, where $\pp$ represents momentum and $\uu_a$ and $\uu_b$
represent the contact manifold normals, we would have a sequence of operations
such as
\begin{align*}
  \pp_1 &= \pp^-\Gamma(\uu_a) \, &\mathrm{with}\quad \inner{\pp_1}{\uu_b} < 0,\\
  \pp_2 &= \pp_1\Gamma(\uu_b)\, &\mathrm{with}\quad \inner{\pp_2}{\uu_a} < 0,\\
  \dots &= \dots \, &\mathrm{with}\quad \dots\notag \\
  \pp^+ &= \pp_{n-1}\Gamma(\uu_b)\, &\mathrm{with}\quad \inner{\pp^+}{\uu_a} \ge 0,\, \inner{\pp^+}{\uu_b} \ge 0.
\end{align*}
The above sequence  generates a solution of the form 
\begin{equation}
  \pp^+ = \pp^-\prod_{i=1}^n\Gamma(\ww_i),
  \label{eq:product}
\end{equation}
where $\prod_{i=1}^{n} x_i$ is short for ``the product $x_A\cdot
x_B\cdots x_n$'' and $\{\ww_i\}$ is a sequence that alternates between
$\uu_a$ and $\uu_b$ as in
$  \{\ww_i\} = \{\uu_a,\uu_b,\uu_a,\uu_b,\dots\}$. 
The solution offered by \eqref{eq:product} is certainly not unique since either
the sequence or the number of terms might vary. For example, a solution
where
$  \{\ww_i\} = \{\uu_b,\uu_a,\uu_b,\uu_a,\dots\} $
might be equally valid, provided that $\pp^+$ is feasible. Similarly, 
since $\Gamma^2(\uu)=I$, prepending the first element an even number of times 
is equivalent to applying the original sequence. Thus, the previous 
alternating sequence is equivalent to 
$  \{\ww_i\} = \{\uu_b,\uu_b,\uu_b,\uu_a,\uu_b,\uu_a,\dots\} $.
Note that we are working with infinite sequences, as we do not want
to assume a priori that a feasible momentum can be found in a finite
number of steps. The question of existence (and thus finiteness of the
length of the sequence) is, contrary to intuition, non-trivial. For a 
discussion and proof of termination for two surfaces, see sec.~\ref{sec:existence}. 

In general, for a set of more than two contact manifold normals 
\begin{equation*}
  S = \left\{ \uu = \frac{D\phi_i(q_*)}{\norm{D\phi_i(q_*)}} \biggm\vert
  \phi_i(q_*) =0, \inner{\pp^-}{\uu}<0 \right\}
\end{equation*}
we will have several choices of $\ww_i \in S$ and a number of terms such
that $\inner{\pp^+}{\uu}\ge0$ for all $\uu\in S.$  With the above in mind we
propose a minimality condition which will reduce the possible mapping sequences
and, in certain cases such as discussed in Sec.~\ref{sec:uniqueness}, will
provide uniqueness.
\begin{definition}[Minimality Condition]
  \label{def:minimal}
  We say that a sequence $W=\{\ww_i\}$ of contact manifold normals is
  \emph{minimal} with respect to an infeasible momentum $\pp$ and Lagrangian
  $L$ if
  the following hold
  \begin{enumerate}
    \item the application of the corresponding sequence of reset maps
      $\{\Gamma(\ww_i)\}$ generates a feasible momentum:
      \begin{equation*}
        \inner{\pp\prod_{i=1}^n\Gamma(\ww_i)}{\ww} \ge 0, 
        \quad \forall\ww\in S
      \end{equation*}
    \item no proper subsequence of $W$ can generate a feasible momentum.
  \end{enumerate}
\end{definition}
We proceed to only consider minimal sequences of reset maps, with two direct
consequences which help us make stronger statements in regards to existence and
uniqueness of feasible solutions to \eqref{eq:two_impact}.  The first, and most
important, is that we do not consider solutions obtained when continuing to
apply reset maps to an already feasible solution.  The second consequence is
that none of the mapping sequences we consider will have consequent members that
are identical, which is guaranteed by the following lemma:
\begin{lemma}
  \label{lemma:doubles}
  Given a sequence $\{\ww_i\}$ which is minimal with respect to some
  momentum $\pp^-$,  we must have that $\ww_j \ne \ww_{j+1}$ for 
  all $j$.\footnote{For clarity, the proof of this and all
      following lemmas can be found in the appendix.}
\end{lemma}
An important consequence of Lemma~\ref{lemma:doubles} is that,
in the case of two contact
manifolds, described by $\uu$ and $\vv$, any minimal sequence of normals has
the form of an alternating sequence of $\uu$ and $\vv$. This fact will be
central to later results.

\begin{example}[Newton's cradle]
Consider the system shown in Fig.~\ref{fig:cradle}. 
We can define manifolds of contact by 
\begin{gather*}
	\phi_1(q) = x_B - x_A - 2r, \\
	\phi_2(q) = x_C - x_B - 2r, \\
	\uu = \D \phi_1(q) = \left[-1, 1, 0\right], \\	
	\vv = \D \phi_2(q) = \left[0, -1, 1\right].
\end{gather*}
The configuration space is three dimensional and the contact manifolds are
two planes. Under the assumption that
all masses are equal $M = m I$, we have that
 \begin{equation*}
 	\Gamma(\uu) = \left[ \begin{array}{rrr} 0 & 1 & 0\\ 1 & 0 & 0\\ 0 & 0 & 1\end{array}\right], \quad 
	\Gamma(\vv) = \left[ \begin{array}{rrr} 1 & 0 & 0\\ 0 & 0 & 1\\ 0 & 1 & 0\end{array}\right].
 \end{equation*}
 It takes three iterations to find a feasible solution, making $\{\uu,\vv,\uu\}$
 and $\{\vv,\uu,\vv\}$ the only minimal sequences with respect to an infeasible
 momentum $\pp^-$: 
 \begin{equation}
   \begin{split}
 	\label{eq:nc_soln}
 	\pp^+ &=  \pp^- \Gamma(\uu) \Gamma(\vv) \Gamma(\uu)
    = \pp^-\Gamma(\vv) \Gamma(\uu) \Gamma(\vv) \\
    & = 
	\pp^-\left[ \begin{array}{rrr} 0 & 0 & 1\\ 0 & 1 & 0\\ 1 & 0 & 0\end{array}\right].
    \end{split}
 \end{equation}
 Note that we obtained the same final result regardless of the order 
 of impacts, which is not expected to hold for general systems.
 For example, if we
 were to choose unequal masses for the balls, the solution would not be,
 in general, unique. This indicates that the uniqueness of the solution is
 related to an interplay between both the geometry of the system and its inertia
 tensor at the time of impact. We will present and
 discuss sufficient conditions for uniqueness in Sec.~\ref{sec:uniqueness}
 and use the results of that
 section to develop the main contribution of the paper, the impact design
 approach in Sec.~\ref{sec:design}.
\end{example}

\subsection{Extension to Inelastic Collisions}
In general, collisions models include a coefficient of restitution, to account
for energy loss during impact.  While the coefficient of restitution is usually
defined as a ratio of collinear impulses\cite{blazejczyk-okolewska_section_1999},
it cannot be used with a propagative model since propagative
models deal with multiple impulse exchanges. 
However, the restitution coefficient 
can be defined
using the energy loss through the impact:
\begin{equation}
  \label{eq:coeff_rest}
  R = \sqrt{1-\frac{\Delta E}{E_p}},
\end{equation}
where $E_p$ is the energy that would have been lost during the collision
through a perfectly plastic impact and $\Delta E$ is the energy lost when
considering the coefficient of restitution $R$.  A coefficient of restitution
of zero will determine a plastic impact while a value of one will generate a
perfectly elastic impact. For all other inelastic impacts we represent the
solution as a convex combination between the plastic and elastic outcomes,
parametrized by $\alpha$:
\begin{equation}
  \label{eq:convex}
  \pp^+ = \alpha\, \pp^+_e + (1-\alpha)\pp^+_p,\quad \alpha\in[0,1],
\end{equation}
where $\pp^+_e$ is the momentum outcome for a perfectly elastic collision and
$\pp^+_p$ is the momentum outcome of a perfectly plastic collision. 
Using \eqref{eq:convex} in the right hand side of \eqref{eq:coeff_rest}, along
with the observation that $E_p = \tfrac{1}{2}\norm{\pp^+_p}^2$, we obtain:
\begin{equation*}
  \norm{\pp^+}^2-\norm{\pp^+_e}^2 = R^2\left(
  \norm{\pp^+_e}^2-\norm{\pp^+_p}^2\right).
\end{equation*}
The values of $\pp^+_e$ and $\pp^+_p$ are given by \eqref{eq:product} and
\eqref{eq:plastic_outcome}, respectively. Since $\pp^+_p$ is
an orthogonal projection of $\pp^-$, and $\pp^+_e$ is conserved
in the direction of $\pp^+_p$, we have that
\begin{equation}
  \norm{\pp^+_p}^2=\inner{\pp^+_p}{\pp^+_e}.
  \label{eq:projection}
\end{equation}
We solve for $\alpha$ and use \eqref{eq:projection} to obtain
\begin{align*}
  \alpha &= \sqrt{1-R^2},\quad R\in [0,1].\\
  \label{eq:inelastic}
\end{align*}

\section{Existence}
\label{sec:existence}
While the previous section presents an overview of the propagative
method for solving simultaneous impacts, it also raises two important questions
regarding the same method: do solutions always exist and, if they do,
are they unique? The current section addresses a special case of existence by
proving that, for a simultaneous impact involving two contact manifolds a minimal
sequence of mappings exists and it is finite. The uniqueness of the corresponding
momentum outcome is
investigated in Sec.~\ref{sec:uniqueness}. 

In what follows we assume that we are dealing with a 
simultaneous elastic impact involving two contact manifolds.
A representation of the contact manifolds at the impact
configuration is given by their normals $\uu$ and $\vv$, which, without any loss
of generality, can be assumed as being of unit length and not collinear:
\begin{subequations}
  \label{eq:normals}
  \begin{align}
  \norm{\uu}&=\norm{\vv}=1,\\
  \uu &\neq \pm\vv.
\end{align}
\end{subequations}
Let $\bbold{T}$ be the intersection
of the two hyperplanes orthogonal to $\uu$ and $\vv$, defined by
\begin{equation*}
  \bbold{T} = \nspace \left\{\uu,\vv\right\}
    =\left\{\pp \in T^*_{q^*}Q|\inner{\pp}{\uu} 
    = \inner{\pp}{\vv} = 0\right\}.
\end{equation*}
Additionally, let $\bbold{N}$ be the plane defined by the two covectors
\begin{equation*}
  \bbold{N} = \spn\{\uu,\vv\}  
  = \left\{ \pp\in T^*_{q^*}Q | \inner{\pp}{\ww}=0,\,\forall\ww\in\bbold{T}\right\}.
  \label{eq:Nspace}
\end{equation*}
Notice that, by definition, $\bbold{T}$ and $\bbold{N}$ are orthogonal and 
complementary. We will make use of these properties
in the following lemmas. We start out by presenting
two lemmas which allow us to equate---through the use of an orthogonal
projection---solutions $\pp \in  T^*_{q^*}Q$
to the lower dimensional  $\pp^\bbold{N} \in \bbold{N}$. The lemmas
provide the link between solving the problem of existence in the lower dimensional
space $\bbold{N}$ and the problem of existence in $T^*_{q^*}Q$: the former
becomes sufficient in order to prove the latter. Thus, the substantial part of
the proof for Theorem~\ref{th:existence} consists of showing existence when 
$\mathrm{dim}(\bbold{N}) = 2$. We do so by using a geometric argument involving the
angle between successive reflection mappings of a covector
$\rr = (\uu+\vv)/\norm{\uu+\vv}$ and a momentum $\pp_0$, which is 
related to the pre-impact momentum $\pp^-$ through an orthogonal projection.
Figure~\ref{fig:proof} illustrates
the geometric argument used in Theorem~\ref{th:existence}.

For the rest of the this section we will use the following notation
to denote the orthogonal projection of a covector onto a set $\bbold{S}$:
\begin{equation*}
  \PP_\bbold{S}(\pp) = \argmin{\pp^*\in\bbold{S}} \norm{\pp-\pp^*}.
\end{equation*}
This notation, along with standard properties of orthogonal projections,
will be heavily used in the following lemmas. First we show that the
feasibility of $\pp$ is equivalent to the
feasibility of its component that can be represented as a linear
combination of $\uu$ and $\vv$.
\begin{lemma}
  \label{lemma:projection}
  Let $\pp, \uu, \vv \in T^*_{q^*}Q$ with $\uu,\vv$ satisfying the restrictions
  in \eqref{eq:normals}.
  Then, $\pp$ is feasible \emph{iff} $\pp^\bbold{N} = \PN(\pp)$
  is feasible.
\end{lemma}
Next, we show that any sequence of reflection transformations will only affect
the component of momentum that is in the span of $\uu$ and $\vv$, leaving the
component orthogonal to that plane unchanged.
\begin{lemma}
  \label{lemma:mappings}
  Given $\pp\in T^*_{q^*}Q$ and a sequence of transformations $\Gamma(\ww_i)$
  with $\ww_i \in \left\{\uu,\vv\right\}$ that map $\pp$ to $\pp_f=\pp\prod_i\Gamma(\ww_i)$,
  we always have that 
  \begin{equation*}
    \pp_f = \PT(\pp) + \PN(\pp)\prod_i\Gamma(\ww_i),
  \end{equation*}
  \end{lemma}
Next, we present a lemma that equates feasibility in $\bbold{N}$ to a
trigonometric condition on the inner product between $\pp$ and a 
covector $\rr$: $\pp$ is feasible \emph{iff} the angle between $\pp$ and $\rr$
is smaller than half the angle between $\uu$ and $\vv$.
\begin{lemma}
  \label{lemma:angle}
  Let $\pp,\uu,\vv,\rr \in \bbold{N}$ such that
  $\rr = \frac{\uu + \vv}{\norm{\uu+\vv}}$ and $\uu\neq\pm\vv$.
  Let $\gamma = \arcsin\inner{\rr}{\uu}$.
  Given the above we
  have that $\pp$ is feasible \emph{iff}
  \begin{equation}
    \inner{\pp}{\rr} \ge \norm{\pp}\cos\gamma.
    \label{eq:angle_lemma}
  \end{equation}
\end{lemma}
Note that for the backwards implication we did not use the fact that $\pp \in
\bbold{N}$, which means that this implication can be easily generalized to more
than two contact manifolds. The same cannot be said about the forward
implication, as counterexamples can be easily found when $\pp \notin
\bbold{N}$: consider the case where $\inner{\pp}{\uu} = \inner{\pp}{\vv} = 0$
but $\norm{\pp}\neq0$.

We use the result of Lemma \ref{lemma:projection} to simplify the problem of existence
of solutions in the higher dimensional cotangent space to an equivalent problem
in a two-dimensional subspace. In particular, since reflection transformations
can only affect the $\PN(\pp)$ component of the momentum, it is enough to
show that, given a finite number of transformations $\Gamma(\ww_i)$, we can
transform $\PN(\pp)$ into a valid momentum $\pp_\bbold{N}\in\bbold{N}$ such
that $\inner{\pp_\bbold{N}}{\uu} \ge 0 \le \inner{\pp_\bbold{N}}{\vv}$. We
can then find the corresponding transformation of the original momentum through
the same sequence of mappings, which, according to Lemma~\ref{lemma:mappings}
is guaranteed to be feasible as well.  Lemma
\ref{lemma:angle} gives us a way to rewrite the feasibility conditions into a
form dependent on the inner product between the momentum and a unit
covector $\rr$ in the cotangent space. Before we show the main result, we
present one last lemma which gives shows that the mapping $\Gamma$ is 
conformal under the kinetic metric:
\begin{lemma}
  \label{lemma:conformal}
  Given a momentum map $\Gamma(\ww)$ and two momenta $\pp_a,\pp_b\in T^*_{q^*}Q$
  we have that 
  \begin{equation}
    \inner{\pp_a\Gamma(\ww)}{\pp_b\Gamma(\ww)} = \inner{\pp_a}{\pp_b}.
  \end{equation}
  In words, $\Gamma(\ww)$ is guaranteed to be a conformal (angle preserving)
  mapping under the kinetic metric.
\end{lemma}
This enables us to formulate and prove the main result:
\begin{theorem}
  \label{th:existence}
  Given a momentum $\pp_0 \in T^*_{q^*}Q$ and two manifolds of impact with
  linearly independent normals $\uu \ne \pm \vv$ at the point of
  simultaneous contact, there exists a minimal sequence $\{\ww_i\}$, as per
  Def.~\ref{def:minimal}, of
  length $n\le \left\lceil\pi/\gamma\right\rceil$ such that
  \begin{equation*}
    \pp_f=\prod_{i=1}^n\Gamma(\ww_i)
  \end{equation*}
  is feasible, where $\gamma$ is the angle defined in Lemma~\ref{lemma:angle}.
\end{theorem}
\begin{IEEEproof}
  We start by considering a momentum $\pp_0^\bbold{N}\in \bbold{N}$.  
  The subspaces $\bbold{T}$ and $\bbold{N}$ are both orthogonal
  and complementary, which makes it possible for us to write
  \begin{equation*}
    \pp_0 = \PT(\pp_0)+\PN(\pp_0).
  \end{equation*}
  We focus our attention on $\pp_0^\bbold{N} = \PN(\pp_0)$, 
  since Lemma~\ref{lemma:projection} guarantees that any feasibility
  results on $\pp_f^\bbold{N}$ extend to $\pp_f$.
  This allows us to work in a two dimensional vector space isomorphic
  to $\reals^2$. Furthermore, Lemma~\ref{lemma:mappings} gives us
  that we can find $\pp_f$ by applying the same sequence of
  mappings to $\pp_0$ as that used when obtaining $\pp_f^\bbold{N}$
  from $\pp_0^\bbold{N}$.

  In what follows we make use of two properties of the momentum map defined in
  \eqref{eq:Gamma}. The first is invertibility. In fact, a mapping
  $\Gamma(\ww)$ is its own inverse, since $\Gamma(\ww)\Gamma(\ww) = I$.  In
  addition, Lemma~\ref{lemma:conformal} shows that $\Gamma(\ww)$ is a conformal
  momentum map under the kinetic energy
  metric: it preserves angles---inner products---according to this metric.

  Now let $\rr_0 = \frac{\uu+\vv}{\norm{\uu+\vv}}$. Using the fact
  that $\Gamma(\ww_i)$ are invertible and taking into consideration
  the result of Lemma~\ref{lemma:angle}, it is sufficient to find 
  a sequence of mappings $\Gamma(\ww_i)$ that maps $\rr$ to
  \begin{equation*}
    \rr_f = \rr_0 \prod_{i=0}^{n} \Gamma(\ww_{i}),
  \end{equation*}
  such that 
  \begin{equation*}
  \inner{\rr_f}{\pp^\bbold{N}_0} \ge \sqrt{\frac{1-\inner{\uu}{\vv}}{2}}\norm{\pp_0^\bbold{N}}.
  \end{equation*}
  Applying the sequence of mappings to $\pp_0$ in reverse order will
  generate a feasible 
  \begin{equation*}
    \pp_f = \pp_0\prod_{i=0}^{n} \Gamma(\ww_{n-i}).
  \end{equation*}
  Given the above, the following has to hold:
  \begin{lemma}
    \label{lemma:induction}
  For any minimal sequence $\{\ww_i\}$ we have that 
  \begin{equation*}
    \inner{\rr_i}{\rr_0} = \cos(2i\gamma),
    \label{eq:rincrease}
  \end{equation*}
  where, as in Lemma~\ref{lemma:angle}, $\gamma = \arcsin\inner{\rr_0}{\uu} =
  \arcsin\inner{\rr_0}{\vv}$ and $\rr_i = \rr_0 \prod_{j=0}^{i} \Gamma(\ww_{j})$.
\end{lemma}

\begin{figure*}
    \centering
    \includegraphics{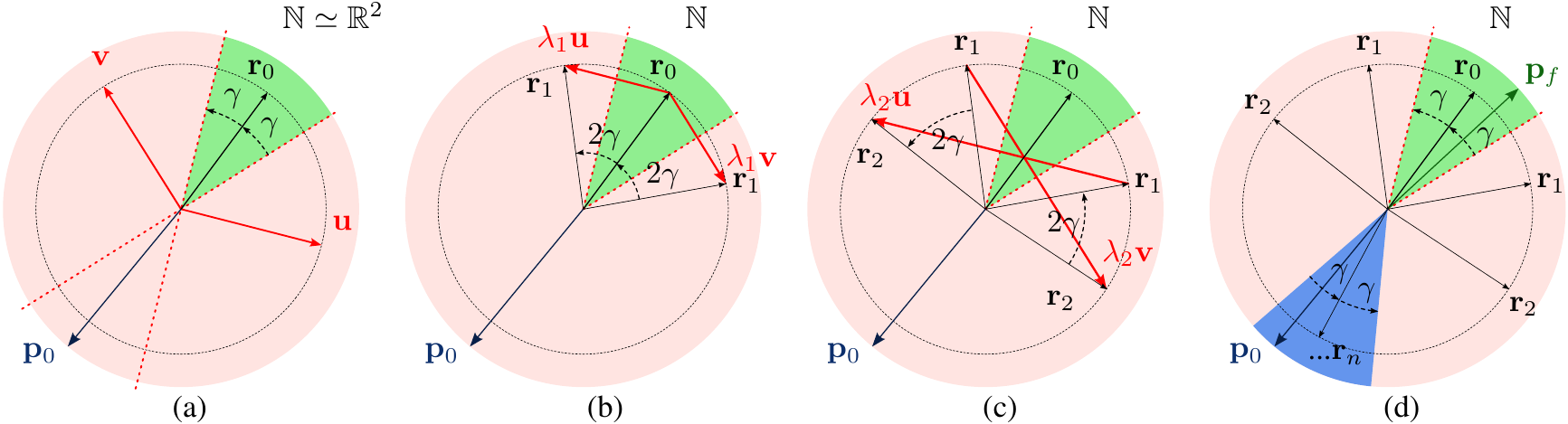}
    \caption{A sketch of the subspace $\bbold{N}$ when the kinetic energy
    metric is identical to the Euclidean metric. The figure gives an 
    interpretation to equations \eqref{eq:angle_increase} and \eqref{eq:R_N}.
    The dotted circle is the intersection of the unit sphere in $T^*_{q^*}Q$
    with $\bbold{N}$ and the dotted lines represent the contact manifolds
    tangents---they are orthogonal to $\uu$ and $\vv$ and represent the axis
    across which the reflection transformations operate. Figure (a) shows the
    two normals $\uu$ and $\vv$, the additional variables $\rr_0$ and $\gamma$
    as well as an initial infeasible momentum $\pp_0$; (b) shows the two
    potential applications of a momentum map to $\rr_0$ and the two
    corresponding values for $\rr_1$; a second sequence of mappings is presented
    in (c)---notice the $2\gamma$ angle increase between $\rr_1$ and $\rr_2$;
    finally, in (d), we have gone through $n$ mappings and obtained an $\rr_n$
    that is within at most $\gamma$ of $\pp_0$. Applying the mappings that
    generated $\rr_n$ to $\pp_0$ in reverse order we obtain $\pp_f$ which is
    guaranteed to be within $\gamma$ of $\rr_0$, and hence feasible.}
    \label{fig:proof}
    \vspace{-1em}
\end{figure*}
  Thus, for any value of $i$ we can calculate $\inner{\rr_i}{\rr_0}=\cos(2i\gamma)$.
  Knowing this angle, the fact that $\rr_i \in \bbold{N}$, and $\norm{\rr_i}=1$,
  we have only two possible $\rr_i$ to satisfy these conditions. One
  of the two covectors is obtained by setting $\ww_0 = \uu$ while
  the other is obtained by setting $\ww_0 = \vv$---see Fig.~\ref{fig:proof}.
  If we account for 
  both these possibilities and consider the set 
  \begin{equation}
    R_N = \left\{\rr_n \left| n \le \left\lceil \frac{\pi}{\gamma}\right\rceil\right.\right\},
    \label{eq:R_N}
  \end{equation}
  we have that, for any $\rr_i \in R_N$, the smallest angle between
  it and any other $\rr_j \in R_N$ is at most $2\gamma$. Formally,
  \begin{equation*}
    \min_i\max_j\inner{\rr_i}{\rr_j} \ge \cos(2\gamma).
  \end{equation*}
  The implication of this is that for any vector $\pp^\bbold{N}$, 
  $\exists i \leq N$ s.t. $\inner{\pp}{\rr_i}\ge\cos(\gamma)$.
  This proves our theorem, since applying the mappings
  used to get $\rr_i$ in reverse order to $\pp_0$ guarantees
  that $\inner{\pp_f^\bbold{N}}{\rr_0}\ge\cos(\gamma)$.
\end{IEEEproof}

The above proof shows that, in the case of two contact manifolds with distinct
normals at the point of contact, a feasible solution can be obtained through a
finite number of successive applications of the reflection mappings defined by
$\Gamma(\uu)$ and $\Gamma(\vv)$. Generalizing to more simultaneous impacts
is complicated by the lack of an alternating structure in ${\ww_i}$ in such cases.
Considering all this, we leave the
proof of such existence an open question to further study.

\section{Uniqueness}
\label{sec:uniqueness}
In this section we present several results that help us determine uniqueness
of simultaneous impacts. In particular, Theorem~\ref{th:orthogonality}
and Theorem~\ref{th:three_stages} guarantee uniqueness of the 
impact results in the cases that the inner product between the two
contact manifold normals takes a value of $0$ or $\pm0.5$, respectively. Also,
as a consequence, if the impact involves a higher number of contact manifolds
which are all pairwise orthogonal at the impact configuration, the result
of applying the reset maps will be unique, regardless of order
of application.

We first show that orthogonality between the normals of the contact manifolds
at the simultaneous impact configurations is both sufficient and necessary
for a unique feasible solution to be found in exactly two steps.
\begin{lemma}[Orthogonality Condition]
  For two contact manifolds described by their normals
  $\uu\neq\pm\vv$, and any infeasible momentum $\pp$ with
  $\inner{\uu}{\pp}\le\inner{\vv}{\pp}<0$, we have that
  \begin{gather}
    \pp_f = \pp\Gamma\left(\uu\right)\Gamma\left(\vv\right) = 
    \pp\Gamma\left(\vv\right)\Gamma\left(\uu\right)\, \textrm{is feasible}\label{eq:commutes}\\
   \emph{iff}\notag\\
   \inner{\uu}{\vv}=0.\label{eq:orthogonal}
 \end{gather}
\end{lemma}
As part of the above result we have that two reset maps commute
if and only if the covectors that generate them are orthogonal. This
helps us show the following important theorem.
\begin{theorem}
  \label{th:orthogonality}
  Given a set of $n$ contact manifolds described by their normals
  $\uu_i$ such that they are all pairwise orthogonal
  \begin{equation*}
    \inner{\uu_i}{\uu_j}=0,\quad\forall i,j,\quad i\neq j,
  \end{equation*}
  and any infeasible momentum $\pp$ with $\inner{\uu_i}{\pp}<0,\,\forall i$,
  we then have that
  \begin{equation}
    \label{eq:cor_prod}
    \pp_f = \pp\prod_i^n \Gamma(\uu_i)
  \end{equation}
  is the same for any ordering of the indices $i$ and the result is feasible.
\end{theorem}
  \begin{IEEEproof}
    It follows directly from Lemma~\ref{th:orthogonality} that, if $\uu_i$
    are pairwise orthogonal then $\Gamma(\uu_i)$ pairwise commute, which
    gives that the order of terms in the product of \eqref{eq:cor_prod} is
    irrelevant.
    
    It remains to show that $\pp_f$ is feasible. We do this
    by showing that the application of a given mapping $\Gamma(\uu_k)$
    affects the inner product between the mapped momentum
    and no other normal but $\uu_k$. Indeed, assume that we have a
    $\pp_k$ and a $\uu_i\neq\uu_k$.
    Then we will have that
    \begin{equation*}
      \begin{split}
      \inner{\pp_k\Gamma(\uu_k)}{\uu_i} &
      = \inner{\pp_k - 2\inner{\pp_k}{\uu_k}\uu_k}{\uu_i}\\
      &= \inner{\pp_k}{\uu_i} - 2\inner{\pp_k}{\uu_k}\inner{\uu_k}{\uu_i}\\
      &= \inner{\pp_k}{\uu_i}.
    \end{split}
    \end{equation*}
    Thus, applying all of the mappings $\Gamma(\uu_i)$ only once, in any
    order, we make sure the sign of the inner product $\inner{\pp_f}{\uu_i}$
    is positive for all normals $\uu_i$. Hence, $\pp_f$ is feasible.
  \end{IEEEproof}

The orthogonality condition in Lemma~\ref{th:orthogonality} was found under the
assumption that the system undergoes two impacts before a feasible exit velocity
is found. Sometimes this might not be the case---e.g. due to design
constraints---and systems where more than two impacts
are needed to find a feasible exit velocity need to be considered. Such a system
is Newton's cradle, where, for equal masses, the 
inner product is $\inner{\uu}{\vv} = -0.5$.
The following
theorem provides sufficient conditions for a three-stage impact to generate
a unique outcome:
\begin{theorem}[Three Stage Impacts]
  \label{th:three_stages}
  Given two contact manifolds described by their normals
  $\uu\neq\pm\vv$
  and $\pp$ that satisfies $\inner{\uu}{\pp}\le\inner{\vv}{\pp}<0$, we have that 
    \begin{gather}
    \pp_f = \pp\Gamma\left(\uu\right)\Gamma\left(\vv\right)\Gamma(\uu) = 
    \pp\Gamma\left(\vv\right)\Gamma\left(\uu\right)\Gamma(\vv) \label{eq:3commutes}\\
    \textrm{is feasible \emph{iff}}\notag\\
    \inner{\uu}{\vv}=-\frac{1}{2}.\label{eq:sixty_degrees}
\end{gather}
\end{theorem}
\begin{IEEEproof}
  Through the use of \eqref{eq:Gamma} and some algebra, the condition in
  \eqref{eq:3commutes} can be shown equivalent to
\begin{equation*}
  \left( 1-4\inner{\uu}{\vv}^2 \right)\left( \inner{\pp}{\uu}\uu - \inner{\pp}{\vv}\vv \right) = 0.
\end{equation*}
Since we assumed that $\uu\neq\pm\vv$, it must be that
\begin{equation*}
  \inner{\uu}{\vv}=\pm\frac{1}{2}.
\end{equation*}
In the case that $\inner{\uu}{\vv}=\frac{1}{2}$, we substitute
back into \eqref{eq:sixty_degrees} to obtain:
\begin{gather*}
  \pp_f = \pp - 2(\vv-\uu)\left(\inner{\pp}{\vv} - \inner{\pp}{\uu}\right),\\
  \inner{\pp_f}{\uu} = 2\inner{\pp}{\uu} - \inner{\pp}{\vv},\\
  \inner{\pp_f}{\vv} = 2\inner{\pp}{\vv} - \inner{\pp}{\uu},
\end{gather*}
which implies that at least some of the infeasible $\pp$ are mapped
to an infeasible $\pp_f$. So $\inner{\uu}{\vv}=\frac{1}{2}$ does not
work, thus showing the forward implication.

On the other hand, if \eqref{eq:sixty_degrees} holds we obtain
\begin{equation*}
  \begin{split}
    \pp_f &= \pp\Gamma\left(\uu\right)\Gamma\left(\vv\right)\Gamma(\uu) \\
    &= \pp - 2\left(\vv-2\inner{\uu}{\vv}\uu\right)\left(\inner{\pp}{\vv}-2\inner{\pp}{\uu}\inner{\uu}{\vv}\right) \\
    &= \pp - 2\left(\vv+\uu\right)\left(\inner{\pp}{\vv}+\inner{\pp}{\uu}\right) \\
    &= \pp\Gamma\left(\vv\right)\Gamma\left(\uu\right)\Gamma(\vv)
  \end{split}
\end{equation*}
Finally, to show that $\pp_f$ is feasible, we calculate
\begin{gather*}
  \begin{split}
  \inner{\pp_f}{\uu} &= \inner{\pp}{\uu}-2\left(1-\inner{\uu}{\vv}\right)\left(\inner{\pp}{\uu}+\inner{\pp}{\vv}\right) \\
                      &= -\inner{\pp}{\vv} > 0,
  \end{split}\\
  \begin{split}
  \inner{\pp_f}{\vv} &= \inner{\pp}{\vv}-2\left(1-\inner{\uu}{\vv}\right)\left(\inner{\pp}{\uu}+\inner{\pp}{\vv}\right) \\
  &= -\inner{\pp}{\uu} > 0.\end{split}
\end{gather*}
\end{IEEEproof}
To illustrate this, consider 
the Newton's cradle system discussed in Sec.~\ref{sec:intro}.
Assuming all three spheres have the same mass $m$, the normals to the
contact manifolds are
\begin{gather*}
  \uu = [-1,1,0]\sqrt{m/2},\\
  \vv = [0,-1,1]\sqrt{m/2}.
\end{gather*}
The dot product between these two normalized covectors is 
\begin{equation*}
  \inner{\uu}{\vv} = m^{-1}\uu\ I\ \vv\tr = -\frac{1}{2},
\end{equation*}
which is consistent with the result of Theorem~\ref{th:three_stages}.
This property of Newton's cradle implies outcome 
uniqueness for any combination of initial momenta of the
spheres
at the moment of simultaneous impact. 

\section{Time-stepping Method for Simultaneous Impact}
\label{sec:discrete}
In the previous sections we have only discussed impact resolution, assuming
that we are already given an impact momentum and impact manifolds. However,
we are interested in simulating a time interval that
contains an impact. Furthermore, our methods for impact resolution presented in
\eqref{eq:one_impact} and \eqref{eq:product} are formulated in terms of the
momentum before and after the impact, the result and proof of
Theorem~\ref{th:existence} is also specified in terms of momentum. We want,
therefore, an integration scheme that works directly with momentum.  The method
of variational
integrators\cite{johnson_scalable_2009,fetecau_nonsmooth_2003,lew_overview_2003,lew_variational_2004}
is ideal in this situation, since, by design, it will conserve a quantity known
as the \emph{discrete momentum}---obtained through the discrete Legendre
transform from a pair of configurations.  The discrete momentum is an
approximation of the continuous momentum at a given time, the error between the
two vanishing in the limit of the time step going to zero.  There are other
advantages to using variational integrators: they are known to preserve the
symplectic form\cite{kane_symplectic-energy-momentum_1999} and conserve the
average energy of the system over a large number of time
steps\cite{fetecau_nonsmooth_2003}.  Finally, recent work has been taking
advantage of variational integrator methods in order to generate optimal
controllers for complex
systems\cite{junge_discrete_2005,leyendecker_discrete_2007,
pekarek_variational_2010,pekarek_variational_2008}. 

Variational integrators are obtained by discretizing the action sum directly:
\begin{multline*}
    L_d(q_i,t_i, q_{i+1},t_{i+1}) \approx \int_{t_i}^{t_{i+1}}L(q,\dot{q},t)\ dt \\
    = \left(t_{i+1} - t_i\right)  
    L\left(\frac{q_i + q_{i+1}}{2}, \frac{q_{i+1} - q_i}{t_{i+1} - t_i},
    \frac{t_i + t_{i+1}}{2}\right).
\end{multline*}
Note that the discrete Lagrangian depends only on configuration variables, and
not on velocity information. There are also several quadrature rules one can 
apply for the discretization.
Here we have used the midpoint rule. 
The discrete equivalent of the Euler-Lagrange equations is
the set of Discrete Euler-Lagrange equations\cite{pekarek_variational_2008}:
\begin{equation}
    \label{eq:DELfull}
    \partl{}{q_k} L_d(q_{k-1}, t_{k-1}, q_k, t_k) + \partl{}{q_k}L_d(q_{k}, t_{k}, q_{k+1}, t_{k+1}) = 0,
\end{equation}
that can be thought of as a mapping from two known configurations $q_{k-1}$
at time $t_{k-1}$ and
$q_k$ at time $t_k$to an unknown configuration $q_{k+1}$ at time $t_{k+1}$.  

A common interpretation of (\ref{eq:DELfull}) is that they enforce the
conservation of \emph{discrete momentum}\cite{fetecau_nonsmooth_2003},
which is defined through the use of
the discrete momentum maps
\begin{align*}
  \FFm(t_a,t_b) &= \partl{}{q_a}L_d(q_a, t_a, q_b, t_b),\\
  \FFp(t_a,t_b) &= \partl{}{q_b}L_d(q_a, t_a, q_b, t_b).
\end{align*}
Using this notation, (\ref{eq:DELfull}) becomes
\begin{equation}
    \label{eq:DEL}
    \FFp(t_{k-1},t_k) + \FFm(t_k,t_{k+1}) = 0,
\end{equation}
which states that the forward momentum $\FFp$ at the end of the $(t_{k-1},t_k)$
interval has to equal the backward momentum $\FFm$ at the beginning of the following
interval, $(t_k,t_{k+1})$. Equation \eqref{eq:DEL} is a discrete equivalent of 
the Euler-Lagrange equations and is known as the DEL (discrete Euler-Lagrange) set
of equations. In case of an impact at time $t_*$, we apply
the update map from \eqref{eq:product} to the discrete momentum, and we solve
\begin{equation}
  \label{eq:DELmap}
  \FFp(t_k,t_*)\prod_i^N\Gamma(\ww_i) + \FFm(t_*,t_k+1) = 0.
\end{equation}
The equations in \eqref{eq:DEL} and \eqref{eq:DELmap} are, for all but the
simplest systems, nontrivial.  
We
use \cite{johnson_scalable_2009} in which the terms in these equations 
are derived using a tree structure and used
in a root finding algorithm. 
This is the method we have used when solving the dynamics away from impact for
the running mechanism described in the following section.

\section{Impact Design}
In this section we apply the uniqueness 
results of Sec.~\ref{sec:uniqueness} to example
systems. We calculate combinations of geometries and configurations
that have unique outcomes for two
mechanical systems: a billiard ball break and a tailed, running biped.
These calculations are done analytically for the billiards and numerically
for the biped.

\subsection{Billiards}
\label{ssec:billiards}
\label{sec:design}\begin{figure}[!b]
  \centering
  \includegraphics{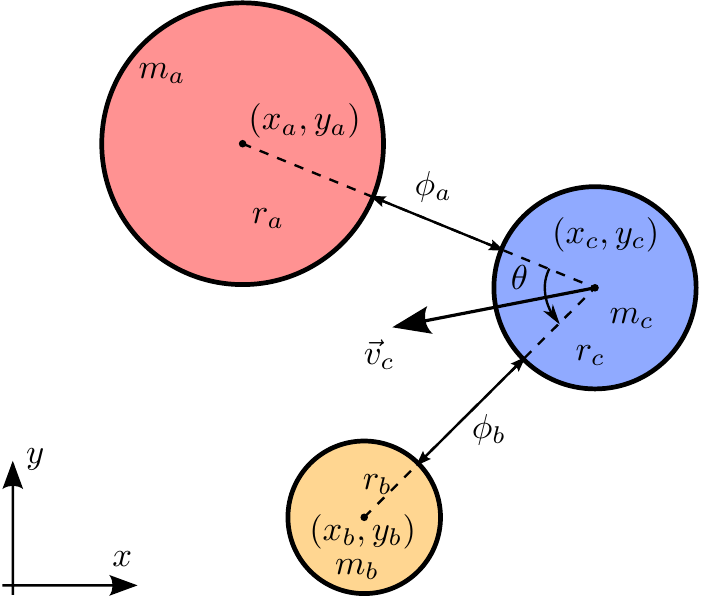}
  \caption{The schematic of a planar mechanical system consisting of
  three billiard balls about to experience simultaneous impact. Ball $a$
  and $b$ are stationary while ball $c$ has an initial velocity $v_c$ such
  that the contact between $a$ and $c$ is simultaneous with the contact
  between $b$ and $c$. The masses of the billiards are proportional to
  the volumes of the respective spheres and all friction effects are ignored.}
  \label{fig:break}
\end{figure}
Consider the general billiard break shown in Fig.~\ref{fig:break}.  Billiard
$c$ acts as the cue ball in this situation, imparting momentum to
the other two billiards through a simultaneous collision.
The configuration vector
for this system is 
\begin{equation*}
  q = [x_a,y_a,x_b,y_b,x_c,y_c]\tr,
\end{equation*}
and the two gap functions are
\begin{gather*}
  \phi_a(q) = \sqrt{(x_a-x_c)^2 + (y_a-y_c)^2} - (r_a + r_c),\\
  \phi_b(q) = \sqrt{(x_b-x_c)^2 + (y_b-y_c)^2} - (r_b + r_c).
\end{gather*}
The two normals at a point where both these functions are zero are
\begin{multline*}
  \D\phi_a(q_*) = \frac{\left[x_c-x_a,\; y_c-y_a,\; 0,\; 0,\; x_a-x_c,\; y_a-y_c\right]}{r_a+r_c}, \\
  \D\phi_b(q_*) = \frac{\left[0,\; 0 ,\; x_c-x_b,\; y_c-y_b ,\; x_b-x_c ,\; y_b-y_c \right]}{r_b+r_c}.
\end{multline*}
\begin{figure}[!t]
  \centering
  \includegraphics{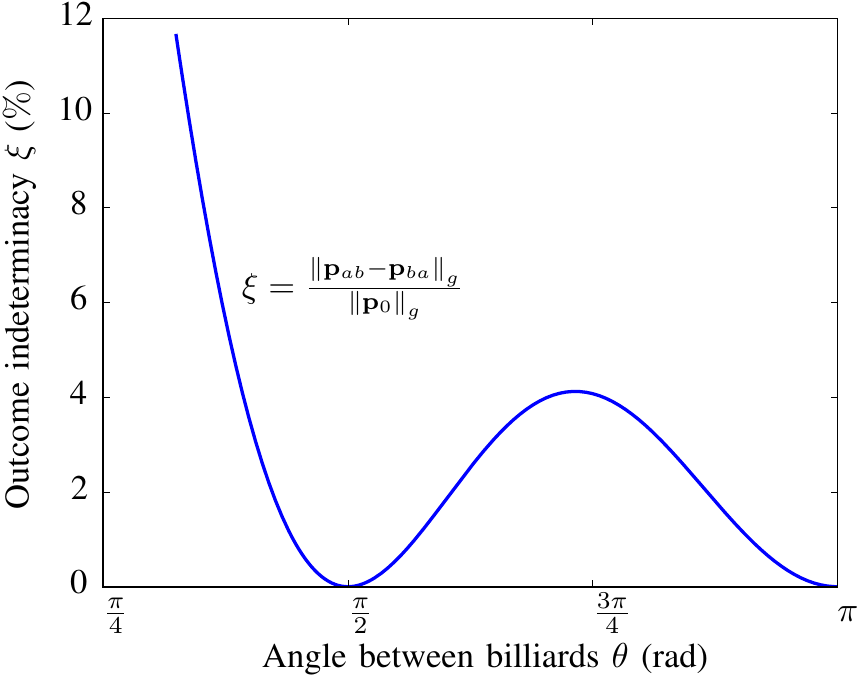}
  \caption{The indeterminacy of the outcome of the billiard break in
  Fig.~\ref{fig:break} as a function of the angle between the billiards.
  As a measure of indeterminacy $\xi$ we looked at the difference in
  momentum between the two possible outcomes under the kinetic
  energy metric and as a percentage of the total energy of the system.}
  \label{fig:billiards}
  \vspace{-0.25in}
\end{figure}

The mass matrix of this system is diagonal and since we are
assuming the impacts to be frictionless and are not expecting any energy in the
rotational modes of the objects, we ignored the moments of inertia when
calculating $M$.  The dot product between the manifolds is
\begin{equation*}
  \inner{\D\phi_a}{\D\phi_b} = \frac{\left(x_a-x_c\right)\left(x_b-x_c\right)
  + \left(y_a-y_c\right)\left( y_b-y_c \right)}{m_c\left( r_a+r_c \right)\left( r_b+r_c \right)}.
\end{equation*}
Since $\phi_a = \phi_b = 0$, we can rewrite this expression in terms
of $\theta$ (see Fig.~\ref{fig:break}) using the law of cosines:
\begin{multline}
  \inner{\D\phi_a}{\D\phi_b} \\= 
    \frac{(x_a-x_b)^2 + (y_a-y_b)^2 - (r_a + r_c)^2 - (r_b + r_c)^2}{2m_c(r_a + r_c)(r_b + r_c)}\\ = 
    \frac{\cos(\theta)}{m_c}.
    \label{eq:billiards}
\end{multline}
Thus, requiring that the impact manifolds be orthogonal under the kinetic
energy metric is, in this case, equivalent to setting $\theta = \pi/2$.  
It is interesting to note that the result in \eqref{eq:billiards} depends
neither on the
masses of the billiards, nor on their radii.
While keeping billiard $c$ in contact with $a$ and $b$,
we varied the angle $\theta$ continuously from the minimum value where $a$ and
$b$ were also in contact---somewhere close to $\pi/6$ in our case---up to
$\pi$. The only billiard with initial velocity was $c$ and $v_c$ was chosen
to point along the bisector of angle $\theta$. As a
measure of indeterminacy we looked at the difference in momentum between the
two possible outcomes under the kinetic energy metric, relative to
the total kinetic energy of the system:
\begin{equation*}
  \xi = \frac{\norm{\pp_{ab}-\pp_{ba}}}{\norm{\pp_0}},
\end{equation*}
where $\pp_0$ is the momentum covector before impact, $\pp_{ab}$ is the
momentum after impact when solving the $a-c$ impact first and $\pp_{ba}$ is the
other option, where we solve for the $b-c$ impact before solving the $a-c$
impact. Figure~\ref{fig:billiards} shows
the values taken by $\xi$ as we varied $\theta$. An expected minimum exists at
$\theta=\pi$, which corresponds to a grazing impact. Somewhat less intuitive
is the minimum at $\pi/2$, which tells us that when the impact configuration
is such that $\theta=\pi/2$ the simulation has no indeterminacy in its solution.

\subsection{TREX}
\begin{figure}[t!]
  \centering
  \includegraphics{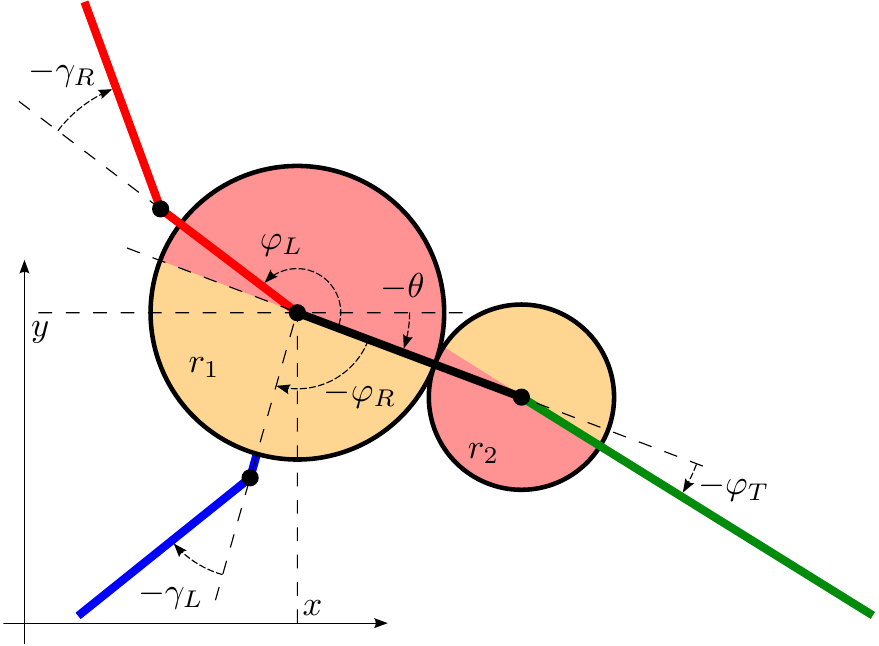}
  \caption{The schematic of the two-dimensional TREX model used in simulation. The
  articulations are at the hips, knees and tail. The degrees of freedom
  consist of the Cartesian coordinates of the main body $x$ and $y$ together
  with all the angles $\varphi$, $\theta$ and $\gamma$, as shown. The floor
  is assumed to be horizontal and positioned at $y=0$.}
  \label{fig:trex}
\end{figure}

The second system we investigated is the tailed running mechanism
(TREX\footnote{
    The name was chosen due to the geometric similarity between
    our mechanism and a commercially available dinosaur
    toy\cite{_blazor_????}.})
in Fig.~\ref{fig:trex}.
The model is inspired by the geometry of several legged locomotors,
such as the RHex\cite{saranli_rhex:_2001}, IMPASS\cite{jeans_impass:_2009} and
several others\cite{eich_versatile_2008,quinn_parallel_2003,lyons_rotational_2005}.
The system consists of a two part body, two articulated
legs and a tail. The two knee joints and the joint between the two parts of the
body are modeled as linear torsional springs and dash pots. The actuators are
located at the hips. The tail is connected rigidly to the posterior body
segment. The configuration variables for this system are
\begin{equation*}
  q = \left[ x,\ y,\ \theta,\ \varphi_L,\ \gamma_L,\ \varphi_R,\ \gamma_R,\ \varphi_T \right]\tr,
\end{equation*}
which represent the position and the orientation of the anterior body, the left
hip and knee angles, the right hip and knee angles and the tail angle. We fixed
the densities of the body and limbs and the section area of the limbs to
reasonable values---the body density is that of water, $1g/cm^3$, the density
of the limbs is that of carbon fiber, $2g/cm^3$, and the cross section of the
limbs was assumed to be $1.25cm^2$---while leaving the leg segment lengths and
radii of the two bodies as design variables. In order to enforce a tapering
shape and reduce the number of variable parameters, we chose the length of the
tail to be a function of the body radii
\begin{equation*}
 L_\textrm{tail} =\frac{r_1^3+r_1^2 r_2+\sqrt{2} \sqrt{r_1^4
  r_2^2+r_1^3 r_2^3}}{r_1^2-r_2^2}-(r_1+r_2),
\end{equation*}
which imposes that the tail length be inversely proportional
with the ratio of the two radii. 

The mechanism drive was generated through external forcing at the hips,
implemented as described in detail in\cite{johnson_scalable_2009,
junge_discrete_2005,leyendecker_discrete_2007,pekarek_variational_2008}.
The values of the forces were chosen by a standard PD controller
with gains $K_p=10^5$ and $K_v=10^4$. Coulomb friction was also added
into the model for this system, in order to facilitate its moving
forward. Friction was implemented by adding an external force in each
independent tangent
direction at the contact point. The coefficients of these forces are found
using the \emph{maximum dissipation principle}\cite{stewart_rigid-body_2000}
which gives rise to a constrained extremization problem that we solved using
standard derivative-free optimization methods at every time step. 

In configurations similar to the one shown in Fig.~\ref{fig:trex} the system
would undergo a simultaneous impact across two manifolds: the tip of the tail
and the tip of one of the legs. For these configurations we can calculate the
angle between the two manifolds, and, according to
Sec.~\ref{sec:uniqueness}, if this angle is $\pi/2$ the indeterminacy of the
outcome will be zero. However, picking a random configuration and set of
parameters such that the two contacts are established will, most likely, return a
non-orthogonal pair of manifolds. This is not hard to imagine, since the two
manifolds themselves depend on the configuration and their dot product is
defined through the kinetic energy metric, which also depends on the inverse of
the mass matrix, and hence on the configuration.

We assume that most of the energy lost through impact comes from dissipation in
the knee joints and that a relatively small amount goes into permanent
deformation of either the robot or the ground.  Thus we model all impacts as
elastic and expect some degree of exponentially decaying chattering. When the
chattering becomes faster than the time step frequency, we assume we have
reached Zeno behavior and consider that impact plastic, in effect taking away
the rest of the energy that would be lost through very high frequency---and
probably not modelable---motion in the dampers. 
\begin{figure}[!t]
      \centering
      \includegraphics{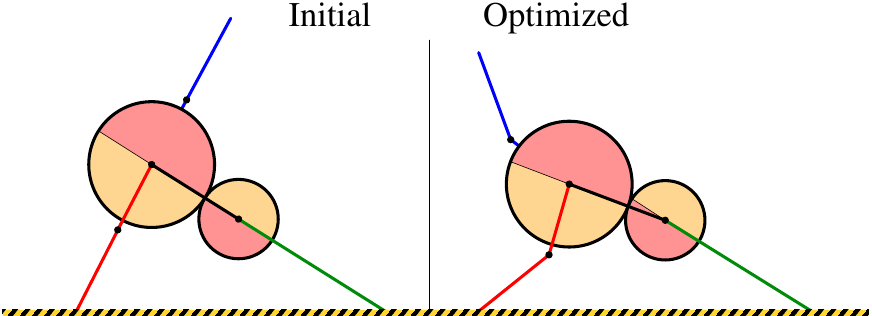}
      \caption{The running robot model (TREX) that we used in order to
      illustrate the results of this paper. Both systems are presented
      in a stance in which double impact is occurring. The model on the right
      has parameters chosen by hand as being reasonable. Its configuration
      and design parameters act as an initial condition for
      an optimization to generate the system and configuration on
      the left. The exact parameters and configuration variables for the
      two systems are also shown (see 
      Table~\ref{tab:trex}).}
  \label{fig:doubletrex}
  \vspace{-0.4in}
\end{figure}

In order to test that our results presented in this paper would have usefulness
in gait and mechanism design, we performed the following simulation. First, we
chose a random configuration of the robot at the time of simultaneous impact, such
that both the tail and the right foot were touching the floor simultaneously.
We made sure that the contact manifolds were not orthogonal under the
kinetic energy metric and labeled the configuration along with the design
parameters to be the \emph{unoptimized system}. Next, we used a classic root
finding algorithm for underconstrained systems in order to find a nearby
configuration and set of parameters for which the dot product between the two
manifolds of contact is zero: this is the \emph{optimized system}. The two
systems are presented in Fig.~\ref{fig:doubletrex} and their parameters
can be seen in Table~\ref{tab:trex}. Furthermore, we also considered a system
with parameters identical to the optimized system but with a \emph{initial stance}:
both the legs were straight at the knee and $180\degree$ out of phase 
at the hips. 
\begin{table*}[!t]
  \caption{Design parameters, configuration variables, and their values
  for both the optimized and unoptimized TREX systems. The largest change
  occurs in $\varphi_R$, which is the right hip angle.}
  \label{tab:trex}
  \begin{center}
  \begin{tabular}{l||*{8}{c}||*{4}{c}||c}
    & \multicolumn{8}{c||}{Configuration variables + Rest angles for torsional springs}
    & \multicolumn{4}{c||}{Design parameters} &  \\
    System &  $x$ & $y$ & $\theta$ & $\varphi_L$ & $\varphi_R$ &
      $\gamma_L$ & $\gamma_R$ & $\varphi_T$ & $r_1$ & $r_2$ &
      $l_u$ & $l_l$ &  $\inner{\uu}{\vv}$ \\
    & $($cm$)$ & $($cm$)$ & $($rad$)$ & $($rad$)$ & $($rad$)$ & $($rad$)$ & $($rad$)$ & $($rad$)$ &
      $($cm$)$ & $($cm$)$  &$($cm$)$ & $($cm$)$ & \\ 
    \hline
    Initial    & 0 & 12.38 & 5.78 & 5.13 & 1.99 & 5.24 & 5.24 & 0 & 7 & 4 & 8 & 10 & $1.65\times10^{-3}$ \\
    Optimized  & 0 & 13.69 & 5.92 & 4.79 & 2.85 & 5.71 & 5.67 & -0.19 & 6.69 & 4.22 & 7.81 & 9.86 &
    $-2.28\times10^{-7}$ 
  \end{tabular}
  \end{center}
\end{table*}

For both the optimized gait and the initial gait we raised the mechanism $17cm$
from the
ground and used the PD controller at the hips to keep the robot in the same
configuration until the moment of impact---at which time the controller applied
torque, spinning the legs counterclockwise in order to create
forward movement. During this first interaction several simultaneous impacts occur,
and we have a choice of which manifold to solve for first: the leg or the
tail. In the first run through we chose the index order for the contact
manifolds based on the $\operatorname{arg\,min}\inner{\ww_i}{\pp_*^+}$. 
\begin{figure}[!t]
  \centering
  \includegraphics{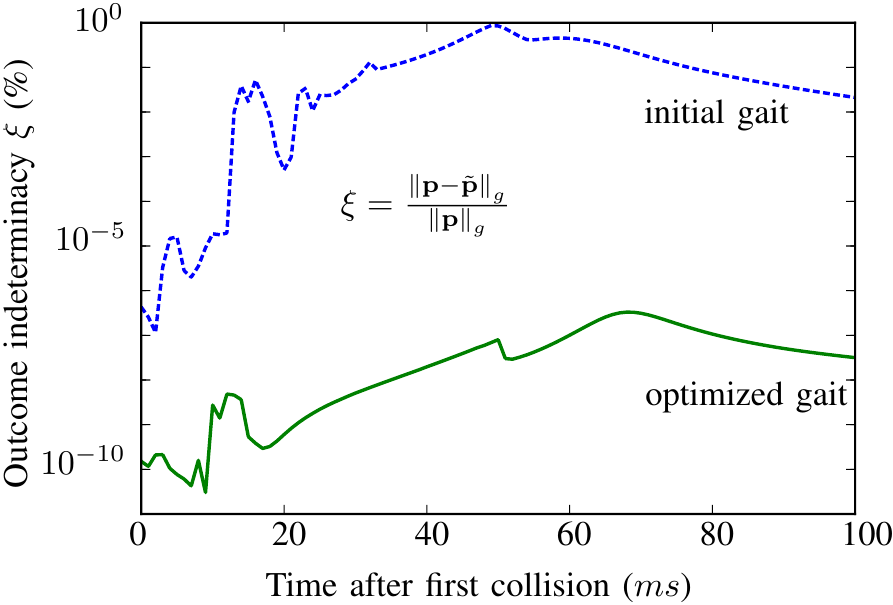}
  \caption{A coordinate-independent measure of the indeterminacy in momentum
  is plotted for two gaits of the optimized mechanism: the optimized gait, and
  the initial gait, in which both legs are straight at the knee and 180$\degree$
  out of phase with each other. The measure of indeterminacy was chosen in an
  analogous way to that in Sec.~\ref{ssec:billiards}. We solved simultaneous
  impacts in each copy of the mechanism by choosing a different order of
  reflections. The 
  normed distance between the momenta of the two mechanism versions at each 
  time point is then normalized by the kinetic energy at that time.} 
  \label{fig:minmax}
  \vspace{-0.3in}
\end{figure}
On the second
run through, however, we reversed this order, effectively using 
$\operatorname{arg\,max}$
instead of $\operatorname{arg\,min}$ for every impact. The  parameters that we considered and
their values for both the unoptimized and optimized system are shown in
Table~\ref{tab:trex}. 

The difference in behavior between the optimized and the initial case
is presented in 
Fig.~\ref{fig:minmax} which shows an indeterminacy measure for each system
based on the normalized uncertainty of the momentum over time. The 
indeterminacy for the optimized gait is more than five orders of magnitude
smaller than that of the initial gait, suggestive of an improvement in
the modelability of the simultaneous impact.

\section{Conclusion}
We have shown how the geometry and configuration of a rigid body system at the
time of simultaneous impact affects its sensitivity to initial conditions, when
a propagative rigid body impact model is used. The measure of sensitivity is obtained
from the inner product between the contact manifold normals at the impact
configuration, and is related to the uniqueness of solutions under the
propagative model---existence of solutions was also shown for the case of two
simultaneous impacts. We optimized two example systems---one analytically, the 
other
numerically---to minimize the sensitivity during a simultaneous impact. Both
optimized and non-optimized versions of the numerical model were simulated using a time 
stepping scheme
based on variational integrators, and significant sensitivity improvement was shown
between the two cases. 

\section*{Acknowledgments}
\addcontentsline{toc}{section}{Acknowledgment}
This material is based upon work supported by the National Science Foundation
under grant IIS-1018167.  Any opinions, findings, and conclusions or
recommendations expressed in this material are those of the authors and do not
necessarily reflect the views of the National Science Foundation.
We would also like to acknowledge Dr. David Pekarek for
the invaluable help and support given through fruitful
conversations relating to problems common of all collision methods, as well
as Dr. Goce Trajcevski for discussions regarding the existence proof.

\bibliographystyle{IEEEtran}
\bibliography{IEEEabrv,tase}

\appendix
\setcounter{lemma}{0}
\begin{lemma}
  Given a sequence $\{\ww_i\}$ which is minimal with respect to some
  momentum $\pp^-$,  we must have that $\ww_j \ne \ww_{j+1}$ for 
  all $j$.
\end{lemma}
\begin{IEEEproof}
  We prove this by reductio ad absurdum: suppose there exists a $j$ such
  that $\ww_j = \ww_{j+1}$. We have then that $\Gamma(\ww_j)\Gamma(\ww_{j+1})=
  \Gamma^2(\ww_j) = I$. This means that we can write:
  \begin{equation*}
    \pp^+=\pp^-\prod_{i=1}^n\Gamma(\ww_i)
    = \pp^-\prod_{i=1}^{j-1}\Gamma(\ww_i)\prod_{k=j+2}^n\Gamma(\ww_k),
  \end{equation*}
  which implies that the subsequence $\{\ww_k\}$ of $\{\ww_i\}$ with
  elements $\ww_j$ and $\ww_{j+1}$ removed also generates a feasible
  momentum. Thus, the sequence $\{\ww_i\}$ cannot be minimal.
\end{IEEEproof}

\begin{lemma}
  Given $\pp\in T^*_{q^*}Q$ and a sequence of transformations $\Gamma(\ww_i)$
  with $\ww_i \in \left\{\uu,\vv\right\}$ that map $\pp$ to $\pp_f=\pp\prod_i\Gamma(\ww_i)$,
  we always have that 
  \begin{equation*}
    \pp_f = \PT(\pp) + \PN(\pp)\prod_i\Gamma(\ww_i),
  \end{equation*}
  \end{lemma}
\begin{IEEEproof}
  It follows from their definitions that $\bbold{T}$ and $\bbold{N}$
  are complementary, such that $\bbold{T}\times\bbold{N}
  = T^*_{q^*}Q.$ Thus, we can write
  \begin{equation*}
    \pp_f=\PT(\pp_f) + \PN(\pp_f).
  \end{equation*}
  Let us look at the first term
  \begin{multline}
    \label{eq:Texpand}
    \PT(\pp_f)=\PT\left[\pp\prod_i\Gamma(\ww_i)\right] \\
    = \PT\left\{\left[\PT(\pp)+\PN(\pp)\right]\prod_i\Gamma(\ww_i)\right\}.
  \end{multline}
  From the definition of $\bbold{T}$ and \eqref{eq:Gamma} we have
  that all transformations that reflect across either the $\uu$ or
  the $\vv$ plane leave vectors in $\bbold{T}$ unchanged. Thus, we can
  say that
  \begin{equation}
    \PT(\pp)\prod_i\Gamma(\ww_i) = \PT(\pp) \in \bbold{T}.
    \label{eq:nullspace}
  \end{equation}
  Similarly, we have that 
  \begin{equation}
    \PN(\pp)\prod_i\Gamma(\ww_i) = \PN\left[\pp\prod_i\Gamma(\ww_i)\right] \in \bbold{N}.
    \label{eq:span}
  \end{equation}
  Substituting \eqref{eq:nullspace} and \eqref{eq:span}  into the right hand
  side of \eqref{eq:Texpand} and using the orthogonality of $\bbold{T}$ and 
  $\bbold{N}$, we obtain
  \begin{equation}
    \PT(\pp_f)=\PT(\pp)
    \label{eq:PTpf}
  \end{equation}
  An identical argument gives the following result for $\PN(\pp_f)$:
  \begin{equation}
    \PN(\pp_f) = \PN(\pp)\prod_i\Gamma(\ww_i).
    \label{eq:PNpf}
  \end{equation}
  The statement of the lemma follows directly from \eqref{eq:PTpf} and 
  \eqref{eq:PNpf}.
\end{IEEEproof}

\begin{lemma}
  Let $\pp,\uu,\vv,\rr \in \bbold{N}$ such that
  $\rr = \frac{\uu + \vv}{\norm{\uu+\vv}}$ and $\uu\ne\pm\vv$.
  Let $\gamma = \arcsin\inner{\rr}{\uu}$.
  Given the above we
  have that $\pp$ is feasible \emph{iff}
  \begin{equation}
    \inner{\pp}{\rr} \ge \norm{\pp}\cos\gamma.
    \label{eq:angle_lemma}
  \end{equation}
\end{lemma}
  \begin{IEEEproof}
  We start by noting that when $\gamma\neq0$---this is true since 
  we assume $\uu$ and $\vv$ are not collinear---several useful relations hold:
  \begin{gather*}
    \inner{\uu}{\vv} = -\cos 2\gamma,\\
      \norm{\uu+\vv} = 2\sin\gamma,
    \end{gather*}
     and
     \begin{equation}
       \inner{\pp}{\rr} = \frac{\inner{\pp}{\uu} + \inner{\pp}{\vv}}{2\sin\gamma}.
       \label{eq:pr}
     \end{equation}

  We first prove the forward implication. Note that the feasibility conditions
  together with \eqref{eq:pr} already give us a lower bound by guaranteeing that
  $\inner{\pp}{\rr}\ge0$.
  Since $\pp \in \spn\{\uu,\vv\}$ we can write
  \begin{equation*}
    \pp = a\uu + b\vv,\quad a,b\in \reals.
  \end{equation*}
  The norm of $\pp$ can be expressed as
  \begin{equation*}
    \norm{\pp} = \sqrt{a^2 - 2ab\cos2\gamma + b^2},
  \end{equation*}
  and the feasibility conditions become
  \begin{gather*}
    a - b\cos 2\gamma \ge 0,\\
    b - a\cos 2\gamma \ge 0,
  \end{gather*}
  which, in turn, imply 
\begin{equation*}
    ab\sin^22\gamma - 2\left(a^2 + b^2\right)\cos2\gamma \ge 0.
\end{equation*}
  Several trigonometric manipulations show that the previous is equivalent to
  \begin{equation*}
    (a+b)^2\sin^2\gamma \ge 
    \cos^2\gamma\left(a^2 - 2ab\cos2\gamma + b^2\right).
  \end{equation*}
  The right hand side can be written in terms of the norm of $\pp$: 
  \begin{equation*}
      (a+b)^2\sin^2\gamma \ge
      \cos^2\gamma\norm{\pp}^2.
  \end{equation*}
We can also write
  \begin{equation*}
    \inner{\pp}{\uu} + \inner{\pp}{\vv} = 2(a + b)\sin^2\gamma,
  \end{equation*}
which, after squaring, gives 
\begin{equation*}
  \frac{\left(\inner{\pp}{\uu} + \inner{\pp}{\vv}\right)^2}{4\sin^2\gamma} = 
    (a+b)^2\sin^2\gamma. 
\end{equation*}
Putting it together, we have
\begin{equation*}
  \inner{\pp}{\rr}^2 \ge \cos^2\gamma\norm{\pp}^2,
\end{equation*}
which proves the forward implication in our lemma.

Now, for the reverse implication, we assume \eqref{eq:angle_lemma} and want to show
that $\pp$ is feasible. Following the last few step of the forward proof, we can
show that \eqref{eq:angle_lemma} is equivalent to 
\begin{equation*}
  \inner{\pp}{\uu} + \inner{\pp}{\vv} \ge \norm{\pp} \sin2\gamma.
\end{equation*}
We want to show that both  $\inner{\pp}{\uu}$ and $\inner{\pp}{\vv}$ are positive.
Let $\uu_t=\uu - \inner{\uu}{\vv}\vv$ be the projection of $\uu$ onto the
plane normal to $\vv$. Using this notation, we write
\begin{equation*}
  \inner{\pp}{\vv} + \inner{\pp}{\uu_t} + \inner{\pp}{\vv}\inner{\uu}{\vv}\ge\norm{\pp}\norm{\uu_t},
\end{equation*}
which, after some algebra, becomes
\begin{equation*}
  \inner{\pp}{\vv}\ge\frac{\norm{\pp}\norm{\uu_t}-\inner{\pp}{\uu_t}}{2\sin^2\gamma}\ge0,
\end{equation*}
where the last inequality holds due to the fact that an inner product is always less than
the product of the vector norms. Using an identical argument we show that
$\inner{\pp}{\uu}\ge0$. The two statements together are equivalent to feasibility.
\end{IEEEproof}

  \begin{lemma}
  For any minimal sequence $\{\ww_i\}$ we have that 
  \begin{equation*}
    \inner{\rr_i}{\rr_0} = \cos(2i\gamma),
    \label{eq:rincrease}
  \end{equation*}
  where, as in Lemma~\ref{lemma:angle}, $\gamma = \arcsin\inner{\rr_0}{\uu} =
  \arcsin\inner{\rr_0}{\vv}$ and $\rr_i = \rr_0 \prod_{j=0}^{i} \Gamma(\ww_{j})$.
\end{lemma}
\begin{IEEEproof}
  We show this using mathematical induction and
  proving that, when $\ww_0 = \uu$ \begin{subequations}
    \label{eq:induction}
  \begin{align}
  \inner{\rr_{k}}{\uu} &= 
    \left\{ \begin{array}{lr}
    \sin[(2k+1)\gamma] & : \ww_{k-1}=\vv \\
    -\sin[(2k-1)\gamma] & : \ww_{k-1}=\uu
    \end{array}\right.\\
  \inner{\rr_{k}}{\vv} &=
    \left\{ \begin{array}{lr}
    -\sin[(2k-1)\gamma] & : \ww_{k-1}=\vv \\
    \sin[(2k+1)\gamma] & : \ww_{k-1}=\uu
    \end{array}\right.\\
  \inner{\rr_k}{\rr_0} &= 
  \cos(2k\gamma)\label{eq:induction_3} 
  \end{align}
  \end{subequations}
   for $\forall k>0$. A symmetric result holds when $\ww_0=\vv$, such
   that \eqref{eq:induction_3} remains unchanged.
   For $k=0$, before any reflections are applied, we have that 
  \begin{align*}
    \inner{\rr_0}{\uu} &= \sin(\gamma),\\
    \inner{\rr_0}{\vv} &= \sin(\gamma),\\
    \inner{\rr_0}{\rr_0} &= \norm{\rr_0} =1.\\
  \end{align*}
  For $k=1$, we make use of the consequence of Lemma~\ref{lemma:doubles} by
  observing that $\{\ww_i\}$ must consist of alternating elements.
  This means that $\ww_1=\uu$ and $\rr_1=\rr_0\Gamma(\uu)$. We calculate
  \begin{align*}
    \inner{\rr_1}{\uu} &= \inner{\rr_0\Gamma(\uu)}{\uu}  = 
      \inner{\rr_0}{\uu} - 2\inner{\rr_0}{\uu}\norm{\uu}^2 \\
      &= \sin(\gamma) - 2\sin(\gamma)\\
      &= -\sin(\gamma),\\
    \inner{\rr_1}{\vv} &= \inner{\rr_0\Gamma(\uu)}{\vv} = 
      \inner{\rr_0}{\vv} - 2\inner{\rr_0}{\uu}\inner{\uu}{\vv} \\
      &= \sin(\gamma) + 2\sin(\gamma)\cos(2\gamma) \\
      &= \sin(3\gamma),\\
    \inner{\rr_1}{\rr_0} &= \inner{\rr_0\Gamma(\uu)}{\rr_0} = 
      \norm{\rr_0} - 2\inner{\rr_0}{\uu}\inner{\uu}{\rr_0}\\
      &=  1 - 2\sin^2(\gamma) \\
      &= \cos(2\gamma),
  \end{align*}
  thus showing the first step of the induction proof when
  $\ww_0=\uu$. The argument is symmetrical for the case when
  $\ww_0=\vv$.
  
  Next, we assume that
  \eqref{eq:induction} holds for $k$ and show that
  it also holds for $k+1$ under the assumption that
  $\ww_{k-1}=\uu$ . We start with
  \begin{multline*}
    \inner{\rr_{k+1}}{\uu} = \inner{\rr_k}{\uu} - 2\inner{\rr_k}{\uu}
    \norm{\uu}   = -\sin[(2k+1)\gamma] \\= -\sin\{[2(k+1)-1]\gamma\}.
  \end{multline*}
  We then show that
  \begin{multline*}
    \inner{\rr_{k+1}}{\vv} = \inner{\rr_k}{\vv} - 2\inner{\rr_k}{\uu}\inner{\uu}{\vv} \\=
      -\sin[(2k-1)\gamma] + 2\sin[(2k+1)\gamma]\cos(2\gamma)\\ = 
        \sin\{[2(k+1)+1]\gamma\}.
  \end{multline*}
  Finally, we have
  \begin{multline*}
    \inner{\rr_{k+1}}{\rr_0} = \inner{\rr_k}{\rr_0} - 2\inner{\rr_k}{\uu}\inner{\uu}{\rr_0} \\=
      \cos(2k\gamma) - 2 \sin[(2k+1)\gamma]\sin(\gamma) \\ = \cos[2(k+1)\gamma] .
  \end{multline*}
  A symmetric argument holds under the complementary assumption
  that $\ww_{k-1} = \vv$. Thus, our inductive
  proof is finished, and we have showed that
  \begin{equation}
    \inner{\rr_n}{\rr_0}=\cos(2n\gamma).
    \label{eq:angle_increase}
  \end{equation}
\end{IEEEproof}

\begin{lemma}
  For two contact manifolds described by their normals
  $\uu\neq\pm\vv$, and any infeasible momentum $\pp$ with
  $\inner{\uu}{\pp}\le\inner{\vv}{\pp}<0$, we have that
  \begin{gather}
    \pp_f = \pp\Gamma\left(\uu\right)\Gamma\left(\vv\right) = 
    \pp\Gamma\left(\vv\right)\Gamma\left(\uu\right)\, \textrm{is feasible}\label{eq:commutes}\\
   \emph{iff}\notag\\
   \inner{\uu}{\vv}=0.\label{eq:orthogonal}
 \end{gather}
\end{lemma}
  \begin{IEEEproof}
    Suppose that \eqref{eq:commutes} holds. We then have that
\begin{equation*}
  \inner{\pp}{\uu}\inner{\uu}{\vv}\vv = 
    \inner{\pp}{\vv}\inner{\uu}{\vv}\uu,\quad \forall\pp\in T^*_{q^*}Q.
  \label{eq:orthoprelim}
\end{equation*}
Since we assume that $\uu \neq \pm \vv$, the only way in which
\eqref{eq:orthoprelim} will hold for any $\pp$ is if
\begin{equation*}
  \inner{\uu}{\vv}=0.
\end{equation*}
Conversely, assuming \eqref{eq:orthogonal}, we can write
\begin{equation*}
  \begin{split}
  \pp_f &=\pp\Gamma(\uu)\Gamma(\vv)  \\
    &= \pp\Gamma(\uu) - 2\inner{\pp\Gamma(\uu)}{\vv}\vv \\
  &= \pp - 2\inner{\pp}{\uu}\uu - 2\inner{\pp}{\vv}\vv 
    + 4\inner{\pp}{\uu}\inner{\uu}{\vv}\vv \\ 
  &= \pp - 2\left(\inner{\pp}{\uu}\uu + \inner{\pp}{\vv}\vv\right) \\ 
  &= \pp\Gamma(\vv)\Gamma(\uu).
\end{split}
\end{equation*}
The symmetry of the result in $\uu$ and $\vv$ assures us of
the commutativity of the $\Gamma(\uu)$ and $\Gamma(\vv)$. The
feasibility of $\pp_f$ is given by the following:
\begin{gather*}
  \inner{\pp_f}{\uu} = \inner{\pp}{\uu} - 2\inner{\pp}{\uu} =
    -\inner{\pp}{\uu} \ge 0, \\
  \inner{\pp_f}{\vv} = \inner{\pp}{\vv} - 2\inner{\pp}{\vv} =
    -\inner{\pp}{\vv} \ge 0,
\end{gather*}
Thus, the statement has been proven.
\end{IEEEproof}

\end{document}